\documentclass[10pt,twocolumn,letterpaper]{article}

\usepackage[pdftex]{graphicx}
\usepackage{subcaption}
\usepackage{float}
\usepackage[justification=raggedright]{caption}	% makes captions ragged right - thanks to Bryce Lobdell
\usepackage{lscape}                                         % Useful for wide tables or figures.
\usepackage{wrapfig}

\usepackage[lined,ruled,linesnumbered]{algorithm2e}
\usepackage{animate} %% convert frames to video

\usepackage{booktabs}                   % Publication quality tables
\usepackage{multirow}

\usepackage{makecell}

\usepackage{paralist}
\usepackage{enumitem}
\setlist[enumerate]{itemsep=0mm}

\usepackage{bm}                          % Make bold, italic math symbols
\usepackage{epsfig}                      % for figures
\usepackage{times}
\usepackage{mathptmx}
\usepackage{mathtools}
\usepackage{amssymb,amsmath}   % Short math guide for LaTeX ftp://ftp.ams.org/pub/tex/doc/amsmath/short-math-guide.pdf

\usepackage{units}
\usepackage{color}

\usepackage{comment}

\usepackage[hyphens]{url}
\usepackage[pagebackref,breaklinks,colorlinks,linkcolor=blue,citecolor=blue,urlcolor=blue,bookmarks=false]{hyperref}
\usepackage{xspace}
\usepackage[table]{xcolor}
\usepackage{setspace}
\usepackage{grfext}
\PrependGraphicsExtensions*{.jpg,.png,.PNG}

\usepackage{gensymb}
\usepackage{soul}
\usepackage{svg}

\usepackage[accsupp]{axessibility} % Improves PDF readability for those with disabilities.

\usepackage[final]{changes}
\def\etal{et~al.\_}			  % and others, and co-workers
\def\eg{e.g.,~}               % for example
\def\ie{i.e.,~}               % that is, in other words
\DeclareMathAlphabet{\altmathcal}{OMS}{cmsy}{m}{n}
\DeclareMathAlphabet{\mathbfit}{OT1}{ptm}{bx}{it}

\newlength\paramargin
\newlength\figmargin

\newlength\secmargin
\newlength\figcapmargin
\newlength\tabcapmargin

\newcommand{\mpage}[2]
{
\begin{minipage}{#1\linewidth}\centering
#2
\end{minipage}
}

\newcommand{\topic}[1]
{
\vspace{1mm}\noindent\textbf{#1}
}

\long\def\ignorethis#1{}
\newbox\jsavebox%

\makeatletter
\newcommand{\providelength}[1]{%
  \@ifundefined{\expandafter\@gobble\string#1}
   {% if #1 is undefined, do \newlength
    \typeout{\string\providelength: making new length \string#1}%
    \newlength{#1}%
   }
   {% else check whether #1 is actually a length
    \sdaau@checkforlength{#1}%
   }%
}

\newcommand{\sdaau@checkforlength}[1]{%
  \edef\sdaau@temp{\expandafter\sdaau@getfive\meaning#1TTTTT$}%
  \ifx\sdaau@temp\sdaau@skipstring
    \typeout{\string\providelength: \string#1 already a length}%
  \else
    \@latex@error
      {\string#1 illegal in \string\providelength}
      {\string#1 is defined, but not with \string\newlength}%
  \fi
}
\def\sdaau@getfive#1#2#3#4#5#6${#1#2#3#4#5}
\edef\sdaau@skipstring{\string\skip}
\makeatother
\usepackage[capitalize]{cleveref}
\crefname{section}{Sec.}{Secs.}
\Crefname{section}{Section}{Sections}
\Crefname{table}{Table}{Tables}
\crefname{table}{Tab.}{Tabs.}

\def\xi{\mathbf{x}_i}

\graphicspath{{figures}, {examples}}

\makeatletter

\def\@fnsymbol#1{\ensuremath{\ifcase#1\or \dagger\or \ddagger\or
\mathsection\or \mathparagraph\or \|\or **\or \dagger\dagger
\or \ddagger\ddagger \else\@ctrerr\fi}}
\makeatother

\graphicspath{{figures/}}

\usepackage[pagenumbers]{cvpr} % To force page numbers, e.g. for an arXiv version
\definecolor{blue}{rgb}{0.4,0.4,0.95}

\begin{document}

\title{
Shape-aware Text-driven Layered Video Editing
}

\author{Yao-Chih Lee \hspace{0.75cm} Ji-Ze Genevieve Jang \hspace{0.75cm} Yi-Ting Chen \hspace{0.75cm} Elizabeth Qiu \hspace{0.75cm} Jia-Bin Huang \\
\vspace{-0.4cm}\\
University of Maryland, College Park\\
\vspace{-0.4cm}\\
\url{https://text-video-edit.github.io}
\vspace{-0.8cm}}

\twocolumn[{
\renewcommand\twocolumn[1][]{#1}
\maketitle

\begin{center}
    \centering
    \frame{\includegraphics[width=0.245\linewidth]{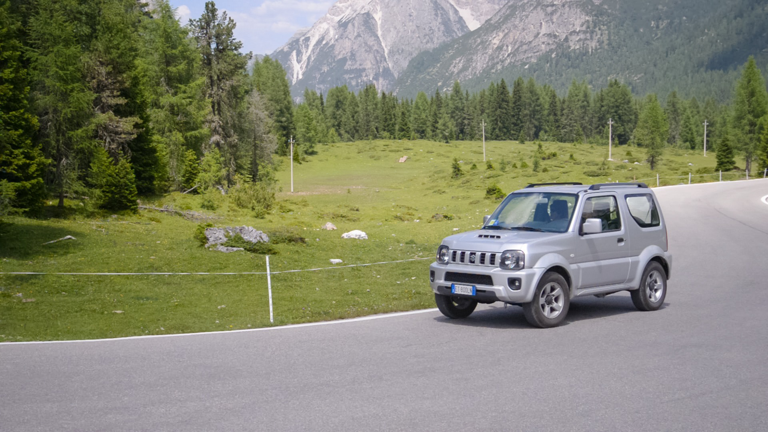}}
    \frame{\includegraphics[width=0.245\linewidth]{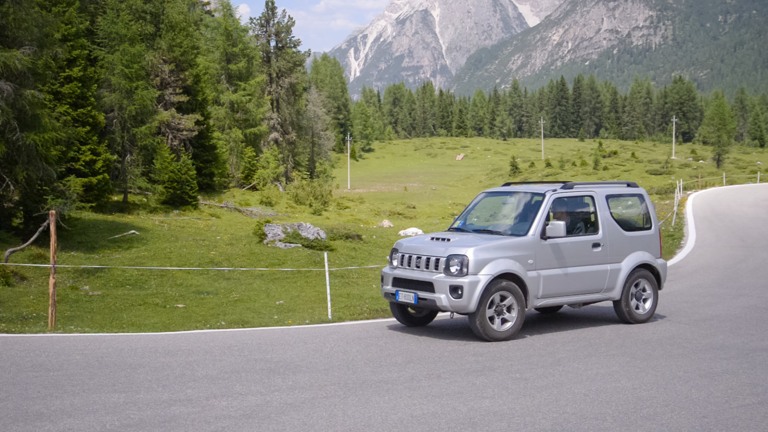}}
    \frame{\includegraphics[width=0.245\linewidth]{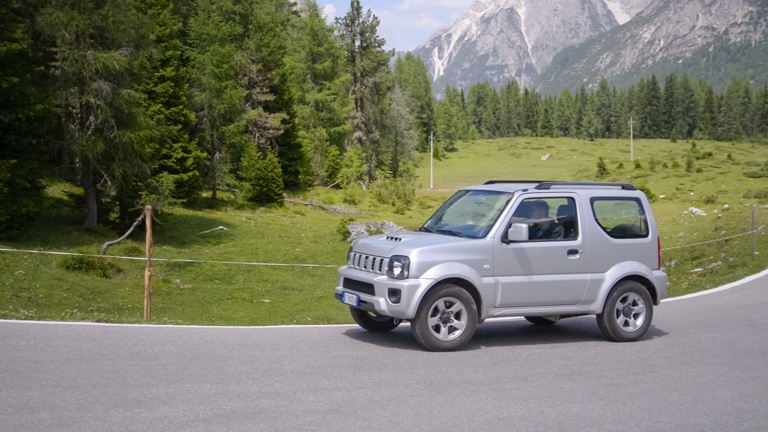}}
    \frame{\includegraphics[width=0.245\linewidth]{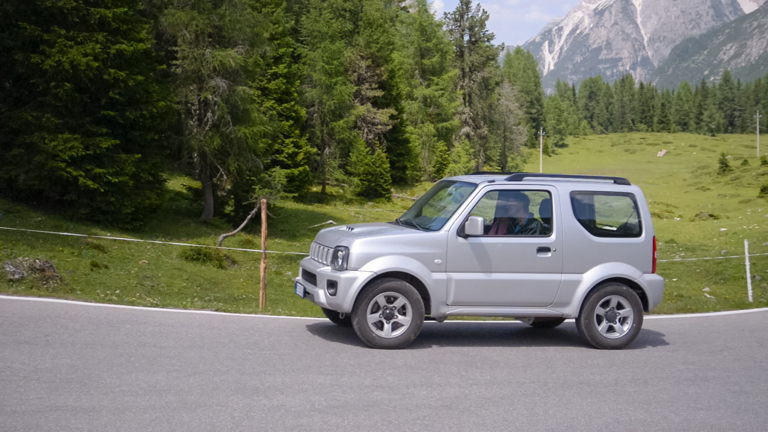}} \\
    \vspace{0.05cm}
    \frame{\includegraphics[width=0.245\linewidth]{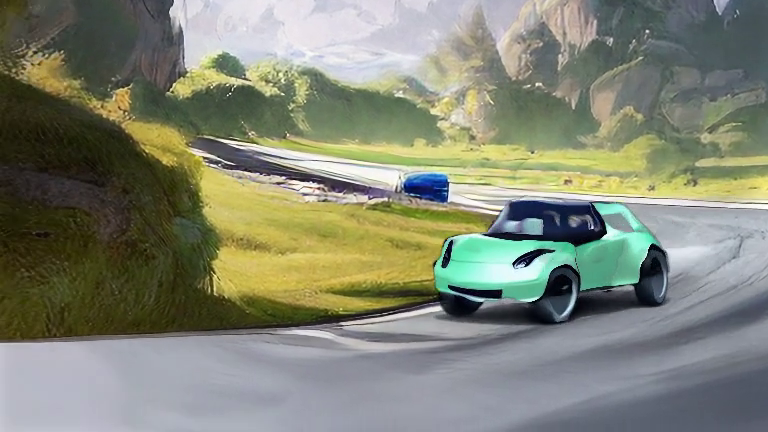}} 
    \frame{\includegraphics[width=0.245\linewidth]{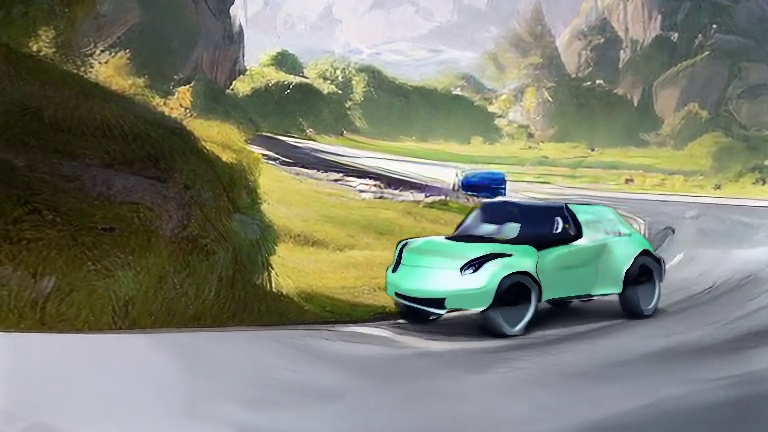}}
    \frame{\includegraphics[width=0.245\linewidth]{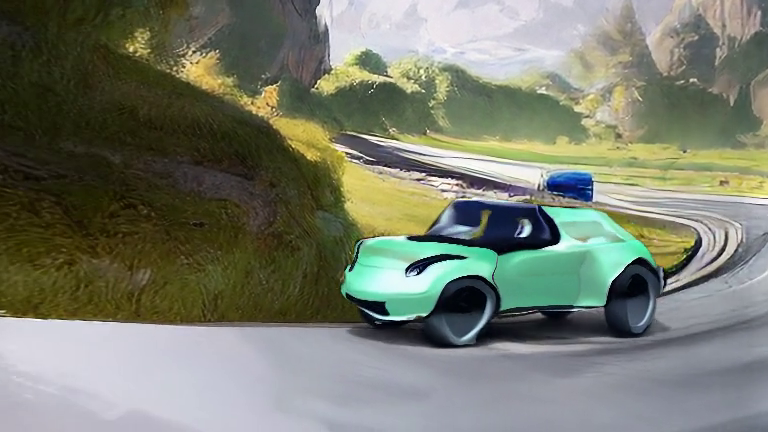}}
    \frame{\includegraphics[width=0.245\linewidth]{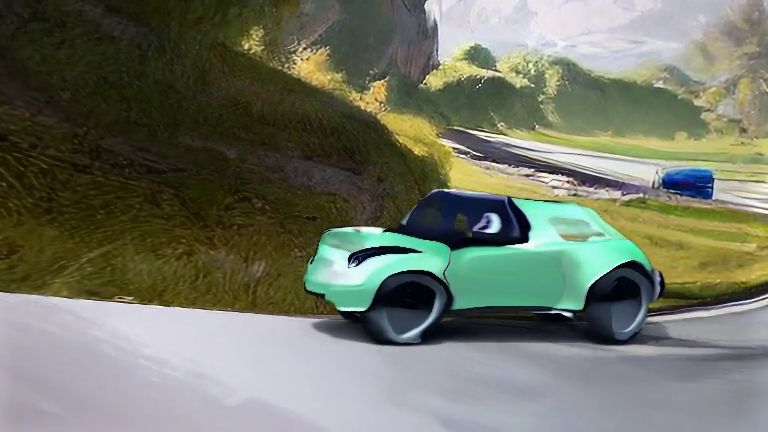}} \\
    \vspace{0.05cm}
    \frame{\includegraphics[width=0.245\linewidth]{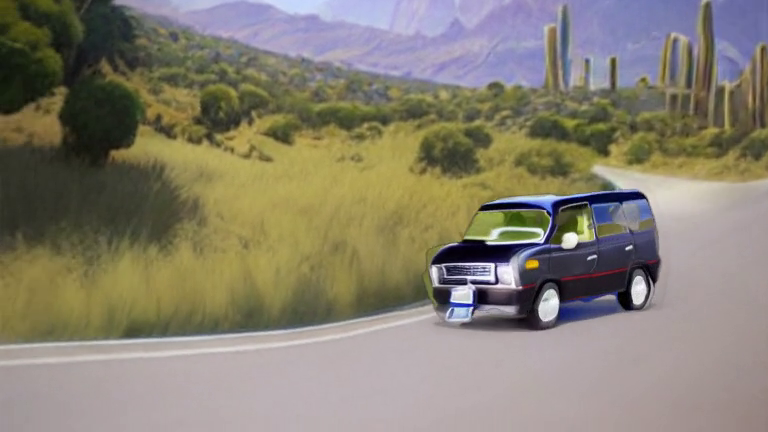}}
    \frame{\includegraphics[width=0.245\linewidth]{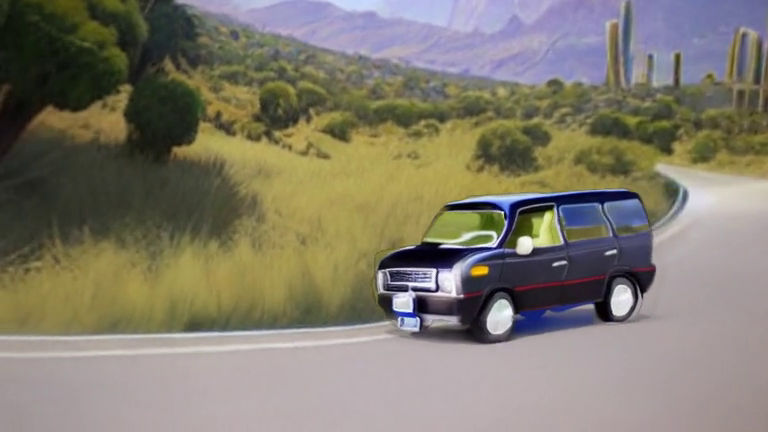}}
    \frame{\includegraphics[width=0.245\linewidth]{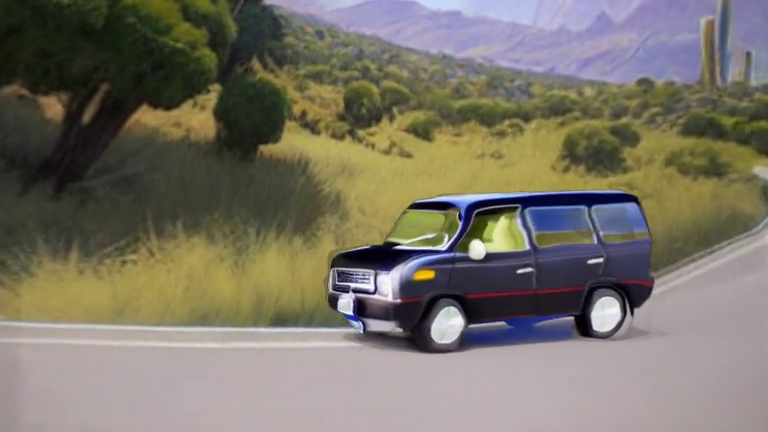}}
    \frame{\includegraphics[width=0.245\linewidth]{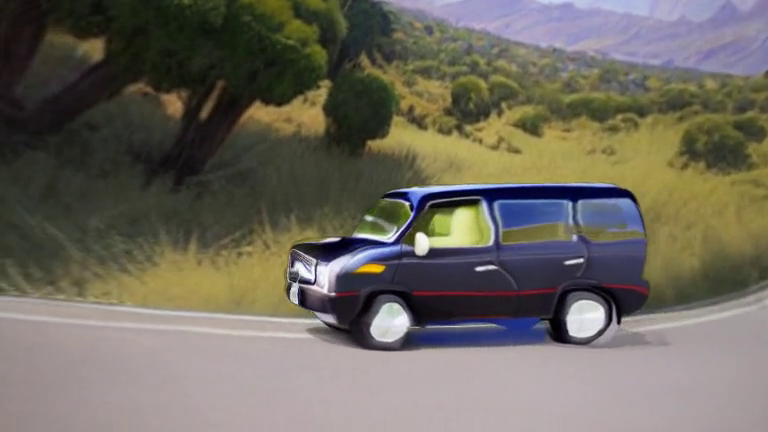}} \\
\captionof{figure}{
\textbf{Shape-aware consistent video editing.} 
Our method enables consistent text-guided video editing with both \emph{appearance} and \emph{shape} changes. 
The top row shows the input frames.  
The second and third rows present editing results from two text prompts: ``\texttt{running sports car}'' and  ``\texttt{running minivan}'', respectively. 
Note that text-driven editing involves \emph{both} texture and structure editing on the foreground object. 
Our method performs consistent edits on sequential frames while preserving the object motion in the input video.
}
    \label{fig:teaser}
\end{center}

}]
\maketitle
\thispagestyle{empty}
\begin{abstract}
Temporal consistency is essential for video editing applications.
Existing work on layered representation of videos allows propagating edits consistently to each frame. 
These methods, however, can only edit object \emph{appearance} rather than object \emph{shape} changes due to the limitation of using a fixed UV mapping field for texture atlas. 
We present a shape-aware, text-driven video editing method to tackle this challenge. 
To handle shape changes in video editing, we first propagate the deformation field between the input and edited keyframe to all frames.
We then leverage a pre-trained text-conditioned diffusion model as guidance for refining shape distortion and completing unseen regions. 
The experimental results demonstrate that our method can achieve shape-aware consistent video editing and compare favorably with the state-of-the-art.
\end{abstract}

%Computational video manipulation is often restricted by the requirement of temporal consistency in contrast to the remarkable success of image manipulation. Previous methods generally rely on video decomposition preprocessing to acquire a unified layer representation (atlas) for editing. The edits on unified layers can be mapped to each frame with consistency. Nevertheless, limited by the fixed mapping fields, the SOTA methods only enable texture changes but no shape editing, which restricts the variety of video manipulation. In this paper, we present a shape-aware, text-guided video editing method to overcome the limitation. Given an input video, we select a keyframe for image editing and propagate it through the precomputed atlas and mapping fields. To unlock the shape changes from the fixed mapping, we propose a deformation formulation by transforming the deformation between the input and edited keyframe to all frames. The per-frame deformation performs consistent shape transformation corresponding to the target keyframe and preserves the object motion in the input video. Furthermore, we leverage a pre-trained diffusion model as guidance to refine the distortion and incomplete edits in an optimization. The experimental results reveal that our method outperforms the SOTA by enabling consistent shape editing corresponding to target prompts. To the best of our knowledge, our method is the first to address the limitation of shape changes in video editing.
\section{Introduction}
\label{sec:intro}
% What is the problem?

% Why is it important? 

% Why is it hard? 

% What other people have done to address it?

% Why do the existing methods not satisfactory?

% What we have done? Start with ``In this paper, "
% - Provide forward references

% What are the specific contributions? Start with ``Our contributions are ..."

\begin{figure*}
    \centering
    (a) Multi-frame editing with frame interpolation~\cite{reda2022film} \\
    \frame{\includegraphics[width=0.245\linewidth]{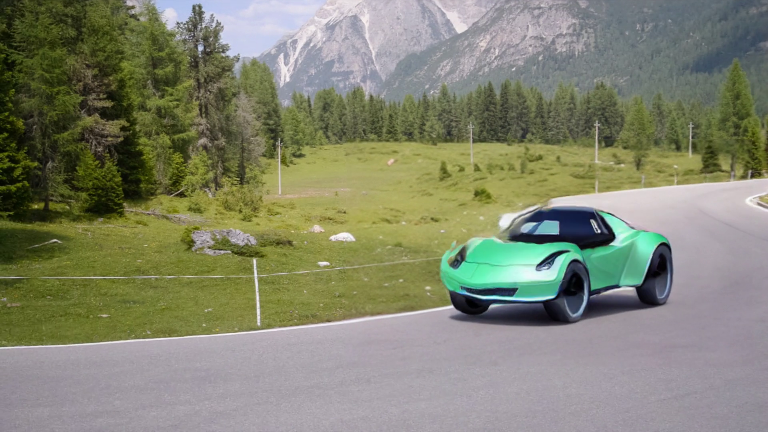}}
    \frame{\includegraphics[width=0.245\linewidth]{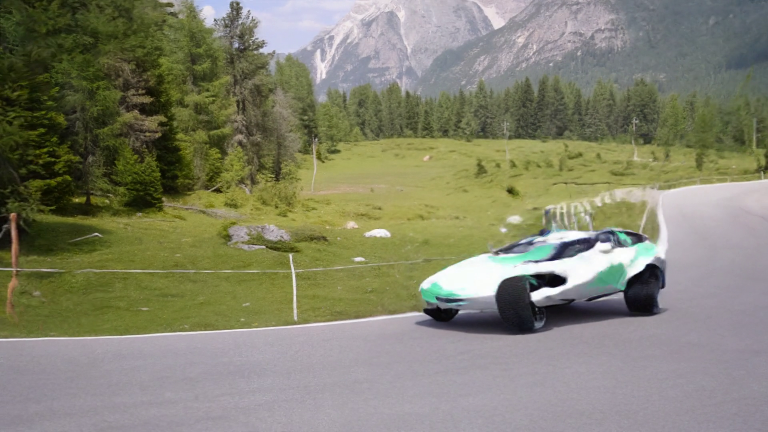}}
    \frame{\includegraphics[width=0.245\linewidth]{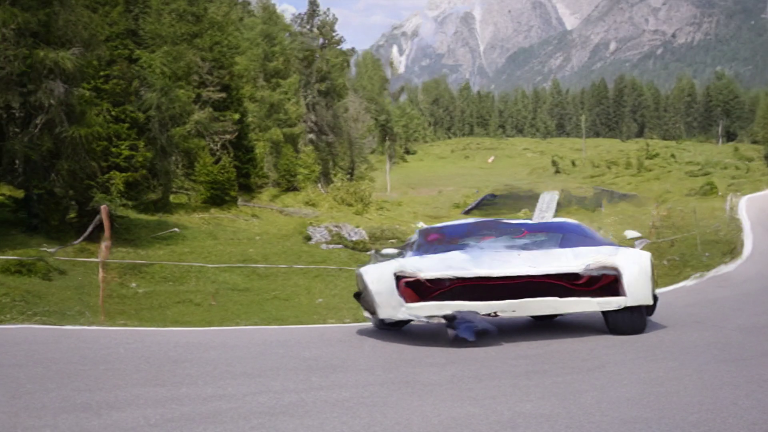}}
    \frame{\includegraphics[width=0.245\linewidth]{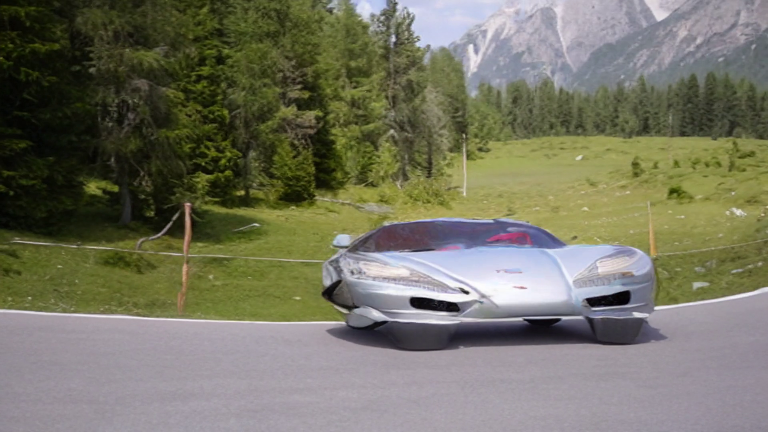}} \\
    (b) Single-frame editing with frame propagation~\cite{jamrivska2019ebsynth} \\
    \frame{\includegraphics[width=0.245\linewidth]{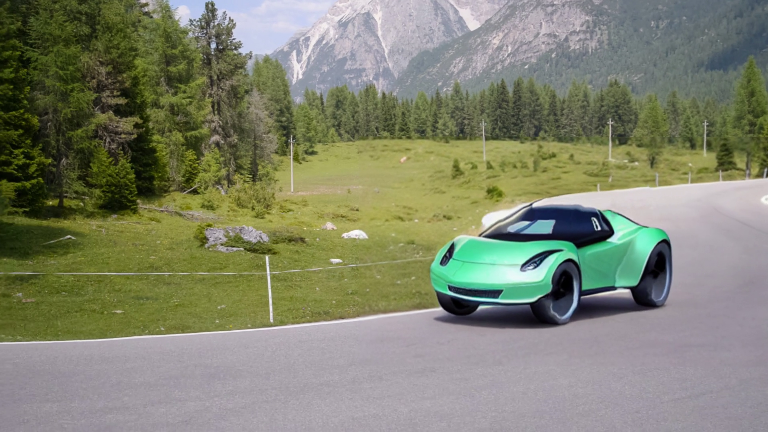}}
    \frame{\includegraphics[width=0.245\linewidth]{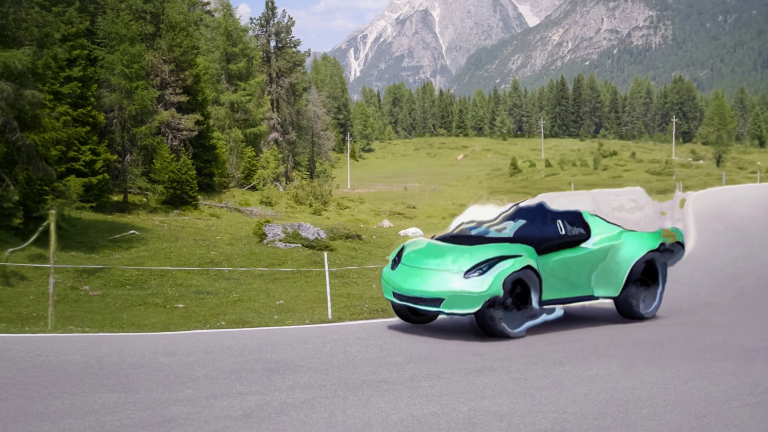}}
    \frame{\includegraphics[width=0.245\linewidth]{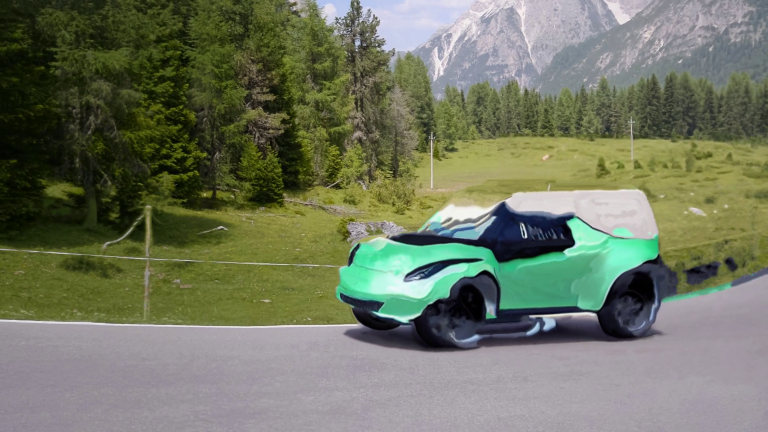}}
    \frame{\includegraphics[width=0.245\linewidth]{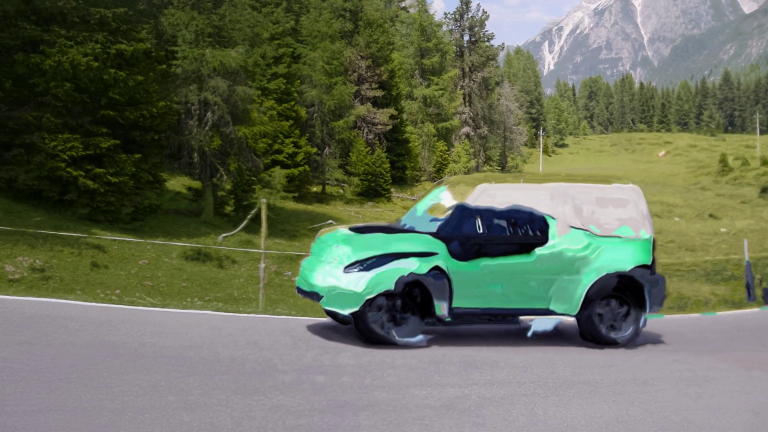}} \\
    (c) Text2LIVE~\cite{bar2022text2live} with prompt ``\texttt{sports car}'' \\
    \frame{\includegraphics[width=0.245\linewidth]{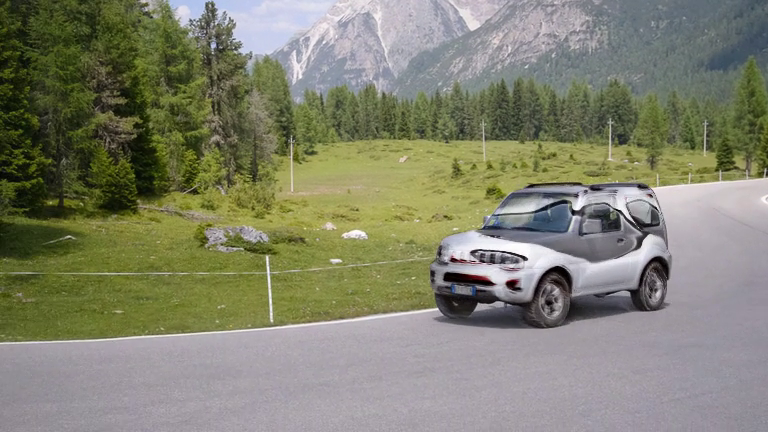}}
    \frame{\includegraphics[width=0.245\linewidth]{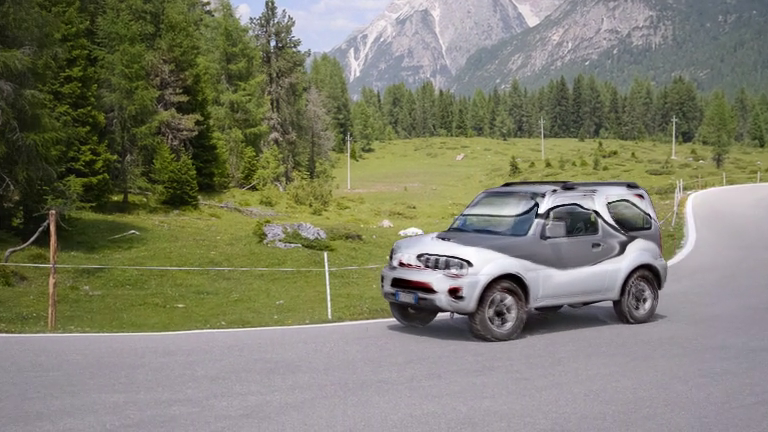}}
    \frame{\includegraphics[width=0.245\linewidth]{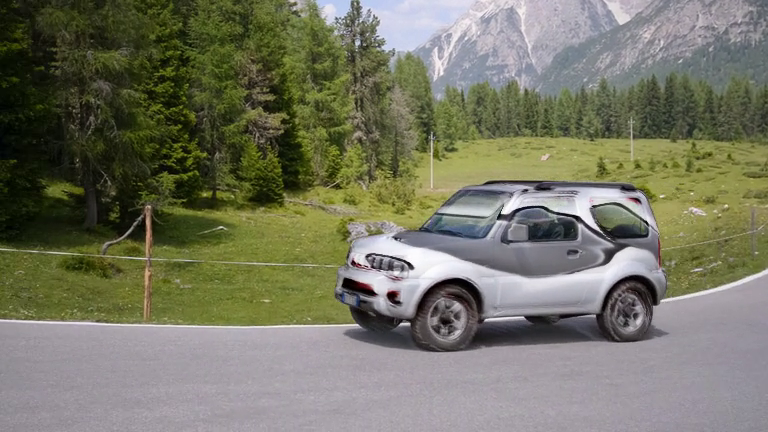}}
    \frame{\includegraphics[width=0.245\linewidth]{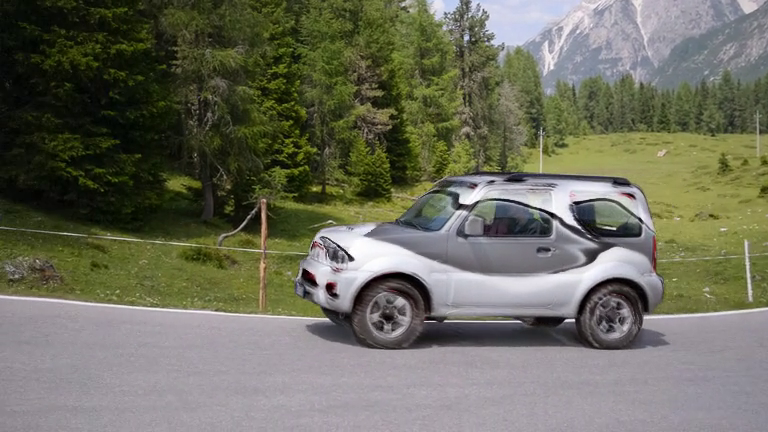}} \\
\caption{\textbf{Limitation of existing work.} 
Compare these results from baseline methods with our ``\texttt{sports car}'' result in Fig.~\ref{fig:teaser}. 
(a) Multiple frames are edited \emph{independently} and interpolated by frame interpolation method~\cite{reda2022film}. 
Such an approach shows realistic per-frame results but suffers from temporal flickering.
(b) Extracting a single keyframe for image editing, the edits are propagated to each frame via~\cite{jamrivska2019ebsynth}. 
The propagated edits are temporally stable. 
However, it yields visible distortions due to the unseen pixels from the keyframe.
% and it uses source-shape frame motion to propagate the new shape edits.
(c) The SOTA Text2LIVE~\cite{bar2022text2live} results demonstrate temporally-consistent appearance editing but remain the source shape ``\texttt{Jeep}'' instead of the target prompt ``\texttt{sports car}'' by using the fixed UV mapping of NLA.}
    \label{fig:limitation_existing_work}
\end{figure*}

%\textcolor{blue}{- Image editings are great
%- Video editings are hard, temporal consistency
%-- propagation (EbSyn) -> artifacts
%-- NLA (texture) -> cannot change shape
%- In this paper, we propose ..
%- Our contributions
%}
\topic{Image editing.}
Recently, image editing~\cite{nam2018text,li2020manigan,rombach2022sd, kawar2022imagic, ramesh2021zero, kim2022diffusionclip} has made tremendous progress, especially those using diffusion models~\cite{rombach2022sd, kawar2022imagic, ramesh2021zero, kim2022diffusionclip}. 
With free-form text prompts, users can obtain photo-realistic edited images without artistic skills or labor-intensive editing. 
However, unlike image editing, video editing is more challenging due to the requirement of temporal consistency.
Independently editing individual frames leads to undesired inconsistent frames, as shown in Fig.~\ref{fig:limitation_existing_work}a. 
A na\"ive way to deal with temporal consistency in video editing is to edit a single frame and then propagate the change to all the other frames. 
Nevertheless, artifacts are presented when there are unseen pixels from the edited frame in the other frames, as shown in Fig.~\ref{fig:limitation_existing_work}b.

%Manual image and video editing are complicated to perform realistic manipulation but also preserve the fidelity of the input image/video. To address the problem, we have witnessed the success of deep learning approaches in image editing~\cite{nam2018text,li2020manigan,rombach2022sd}. Users can type in a text prompt of the desired editing effect to manipulate a source image with recent Vision-Language models~\cite{radford2021clip}.
%Similarly, computational video editing is essential to save users' efforts from intricate frame-by-frame manual manipulation. Nonetheless, video editing is more challenging than image editing due to the requirement of temporal consistency. As shown in Fig.~\ref{fig:limitation_existing_work}a, the inconsistency can be easily observed if an image-based manipulation model independently edits the frames.

%For the sake of temporal consistency, a naive approach is to manipulate a single frame of a video with an image editor and then propagate the edited pixels to all other frames by the motions in the video. Nevertheless, this approach would yield many artifacts since the unseen pixels from the single edited frame in different frames could be inconsistent with the edited frame (as shown in Fig.~\ref{fig:limitation_existing_work}b).

\topic{Video editing and their limitations.}
For consistent video editing, \emph{Neural Layered Atlas} (NLA)~\cite{kasten2021nla} decomposes a video into unified appearance layers \emph{atlas}. 
The layered decomposition helps consistently propagate the user edit to individual frames with per-frame UV sampling association. 
Based on NLA, Text2LIVE~\cite{bar2022text2live} performs text-driven editing on atlases with the guidance of the Vision-Language model, CLIP~\cite{radford2021clip}. 
Although Text2LIVE~\cite{bar2022text2live} makes video editing easier with a text prompt, it can only achieve \emph{appearance manipulation} due to the use of fixed-shape associated UV sampling. 
Since per-frame UV sampling gathers information on motion and shape transformation in each frame to learn the pixel mapping from the atlas, shape editing is not feasible, as shown in Fig.~\ref{fig:limitation_existing_work}c.

%For consistent video editing, \emph{Neural Layered Atlas} (NLA)~\cite{kasten2021nla} is proposed to decompose a video into unified appearance layers, termed \emph{atlas}. With the associated per-frame UV sampling, the user-manual editing on an atlas could be spread to each frame consistently. 

%Based on NLA, Text2LIVE~\cite{bar2022text2live} performs automatic text-driven editing on the atlas with the guidance of the Vision-Language model, CLIP~\cite{radford2021clip}. However, both NLA and Text2LIVE enable \emph{appearance-only} manipulation due to the limitation of associated UV sampling. Specifically, the per-frame UV sampling stores each frame's motion and shape transformation to map from the atlas. Thus, the shape editing is disabled by the fixed UV mapping (Fig.~\ref{fig:limitation_existing_work}c).

\topic{Our work.} 
In this paper, we propose a \emph{shape-aware} text-guided video editing approach. 
The core idea in our work lies in a novel UV map deformation formulation. 
With a selected keyframe and target text prompt, we first generate an edited frame by image-based editing tool (\eg, Stable Diffusion~\cite{rombach2022sd}).
We then perform pixel-wise alignment between the input and edited keyframe pair through a semantic correspondence method~\cite{truong2022probabilistic}.
The correspondence specifies the deformation between the input-edited pair at the keyframe. 
According to the correspondence, the shape and appearance change can then be mapped back to the atlas space. 
We can thus obtain per-frame deformation by sampling the deformation from the atlas to the original UV maps. 
While this method helps with shape-aware editing, it is insufficient due to unseen pixels in the edited keyframe.
We tackle this by further optimizing the atlas texture and the deformation using a pretrained diffusion model by adopting the gradient update procedure described in DreamFusion~\cite{poole2022dreamfusion}. 
Through the atlas optimization, we achieve consistent \emph{shape} and \emph{appearance} editing, even in challenging cases where the moving object undergoes 3D transformation (Fig.~\ref{fig:teaser}).

\topic{Our contributions.} 
\begin{itemize}
    \item We extend the capability of existing video editing methods to enable shape-aware editing.
    \item We present a deformation formulation for frame-dependent shape deformation to handle target shape edits.
    \item We demonstrate the use of a pre-trained diffusion model for guiding atlas completion in layered video representation.
    % The shape and appearance parameters in the atlas are optimized by a pre-trained diffusion model to complete unseen regions.
\end{itemize}

\section{Related Work}
\label{sec:related}

% Relationship examples:
% Similar:
% - Our work also adopt X ...
% - We address similar challenges.
% Different:
% - Our work differs in X ...
% - In contrast, we tackle ...

\topic{Text-driven image synthesis and editing.} 
Recent years have witnessed impressive progress in text-guided image synthesis and manipulation using GANs~\cite{reed2016generative,zhang2017stackgan,xu2018attngan,li2019object,li2020manigan,xia2021tedigan,ramesh2021dalle,liao2022text}. 
On text-to-image \emph{generation}, DALL-E~\cite{ramesh2021dalle} first demonstrates the benefits of training text-to-image models using a massive image-text dataset. 
Most recent text-to-image generators~\cite{liu2021fusedream,crowson2022vqgan} use a pre-trained CLIP~\cite{radford2021clip} as the guidance.
On text-to-image \emph{manipulation/editing}, recent methods also take advantage of the pretrained CLIP embedding for text-driven editing~\cite{patashnik2021styleclip,gal2022stylegannada,xu2022predict}.
These methods either pretrain the model with CLIP embedding as inputs or use a test-time optimization approach~\cite{frans2021clipdraw,kwon2022clipstyler,bar2022text2live}.

% StyleCLIP~\cite{patashnik2021styleclip} exploits pre-trained StyleGAN and CLIP to show descent appearance and shape editing.
% On the other hand, the test-time optimization approach~\cite{frans2021clipdraw,kwon2022clipstyler,bar2022text2live} performs training for each testing instance by leveraging CLIP as guidance without pre-training.

Recently, diffusion models~\cite{ho2020denoising,song2020denoising,dhariwal2021diffusion} have shown remarkable success in both text-guided image generation~\cite{nichol2022glide,rombach2022sd,saharia2022imagen,ruiz2022dreambooth,balaji2022ediffi} and editing~\cite{nichol2022glide,rombach2022sd,hertz2022prompt} tasks. 
Stable Diffusion~\cite{rombach2022sd} performs a denoising diffusion process in a latent space and achieves high-resolution text-to-image generation and image-to-image translation results.
In particular, the release of the model trained on large-scale text-image pair dataset~\cite{schuhmann2022laion5b} facilitates various creative applications from artists and practitioners in the community. %\cite{schuhmann2022laion5b}
% to enable a huge variety of synthesis and editing.
% Relationship
Our work leverages the state-of-the-art text-to-image model, Stable Diffusion~\cite{rombach2022sd}, and extends its semantic image editing capability to consistent video editing.

\topic{Video generation.}
Building upon the success of photorealistic (text-driven) image generation, recent work has shown impressive results on video generation, with a focus on generating long video~\cite{ge2022long,brooks2022generating,skorokhodov2022stylegan,yu2022generating} and videos from free-form text prompts~\cite{singer2022make,ho2022imagen,villegas2022phenaki}.
Unlike video \emph{generation} methods, our work differs in that we perform text-driven video \emph{editing} for real videos.

\topic{Video editing.}
In contrast to the breakthrough of image editing, video editing methods are faced with two core challenges: 1) temporal consistency and 2) computational complexity of the additional dimension. 
To attain temporally consistent editing effects, EbSynth~\cite{jamrivska2019ebsynth} utilizes keyframes and propagates the edits to the entire video with optical flows computed from consecutive frames.
Such flow-based techniques have been applied in other tasks such as video synthesis~\cite{bhat2004flow}, video completion~\cite{huang2016temporally,gao2020flow,li2022towards}, and blind video consistency~\cite{lai2018learning,lei2022deep}.
Several studies address temporal inconsistency in the latent space via GAN inversion ~\cite{liu2022deepfacevideoediting,xu2022temporally,xia2022gan}.
However, current GAN-based models can only model datasets with limited diversity (e.g., portrait or animal faces).
Another line of approaches~\cite{lin2017layerbuilder,lu2020retiming,kasten2021nla,lu2021omnimatte,ye2022sprites} decomposes a video into unified layer representation for consistent editing. 
Neural Layered Atlas (NLA)~\cite{kasten2021nla} performs test-time optimization on a given input video to learn the canonical appearance layer and per-frame UV mapping using video reconstruction loss. 
With layer decomposition, one can use text-driven image editing techniques to the unified layers to consistently broadcast the edits to each frame. 
The work most relevant to ours is Text2LIVE~\cite{bar2022text2live} and Loeschcke~\etal~\cite{loeschcke2022text}.
Both methods build upon NLA to perform text-driven editing on the learned atlases. 
A pre-trained CLIP is used for each input video to guide the atlas editing via a test-time optimization framework. 
Yet, limited by the formulation of NLA, they only allow \emph{appearance} edits due to the fixed UV mapping from the atlas to frames. 
The mapping fields store the original shape information in each frame so that the fixed UV mapping restricts the freedom of \emph{shape editing} in~\cite{kasten2021nla,bar2022text2live,loeschcke2022text}. 
Our work also builds upon NLA for achieving temporally consistent video editing. 
In contrast to existing methods~\cite{bar2022text2live,loeschcke2022text}, we extend the capability of text-driven editing to enable shape editing.
\def\D{\altmathcal{D}}  % deformation map
\def\I{\altmathcal{I}}  % image/frame
\def\m{\altmathcal{M}}  % mask
\def\S{\altmathcal{S}}  % shape
\def\uv{\altmathcal{W}}  % UV map
\def\a{\altmathcal{\alpha}}  % alpha map
\def\res{\altmathcal{R}}

% Vectors
\def\b{\mathbfit{b}}
\def\c{\mathbfit{c}}
\def\d{\mathbfit{d}}
\def\o{\mathbfit{o}}
\def\p{\mathbfit{p}}
\def\t{\mathbfit{t}}
\def\x{\mathbfit{x}}
\def\z{\mathbfit{z}} % latent
\def\e{\epsilon}

% Matrices
\def\K{\mathbfit{K}}
\def\R{\mathbb{R}} % real space
\def\M{\mathbfit{M}} % transformation matrix for shift vectors

% Functions
\def\ang{\phi}
\def\dehom{\mu}
\def\proj{\pi}
\def\sigmoid{S}
\def\vis{\nu}
\def\r{\mathbfit{r}}
\def\T{\mathbfit{T}} % deformation transformation
\def\L{\altmathcal{L}} % loss function

% Bracketed
\def\bp{(\p\!)} % (p)
\def\bt{(t\!)} % (p)
\def\bx{(\x\neg)} % (x)

% Subsets
\def\ok{\o_{\neg k}}
\def\tk{\t_{\neg k}}
\def\wk{w_{\neg k}}
\def\xi{\x_{\neg i}}
\def\zk{\z_{\neg k}}
\def\Kk{\K_{\neg k}}
\def\Rk{\R_{\neg k}}

% Helpers
\def\ng{\hspace{-0.1mm}}
\def\neg{\hspace{-0.2mm}}
\def\pos{\hspace{0.2mm}}

% Create \overrightharpoon, as a replacement for \overrightvector.
\makeatletter
\newcommand*\MY@rightharpoonupfill@{%
    \arrowfill@\relbar\relbar\rightharpoonup
}
\newcommand*\overrightharpoon{%
    \mathpalette{\overarrow@\MY@rightharpoonupfill@}%
}
\makeatother

% A scalable sum symbol. Use like this: \nsum[0.8]
\newlength{\depthofsumsign}
\setlength{\depthofsumsign}{\depthof{$\sum$}}
\newcommand{\nsum}[1][1.4]{% only for \displaystyle
    \mathop{%
        \raisebox
            {-#1\depthofsumsign+1\depthofsumsign}
            {\scalebox
                {#1}
                {$\displaystyle\sum$}%
            }
    }
}

%%%%%%%%%%
%%%%%%%%%%
\section{Method}
\label{sec:method}
\begin{figure*}
    \centering
    \includegraphics[width=\linewidth]{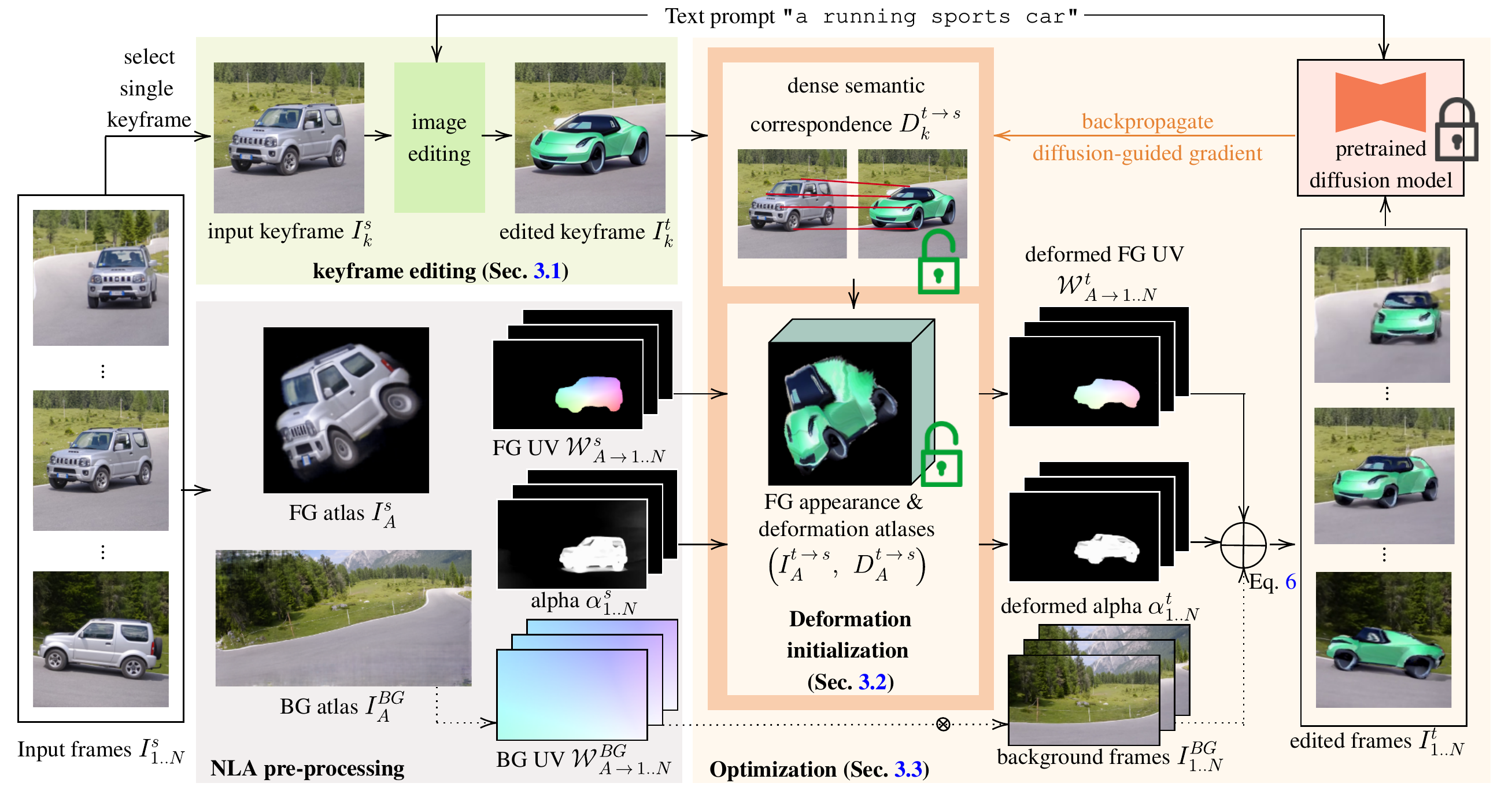}
\caption{\textbf{Method overview.} 
Given an input video and a target edit text prompt, our method first bases on a pre-trained NLA~\cite{kasten2021nla} to decompose the video into unified atlases with the associated per-frame UV mapping. 
Aside from video decomposition, we use the text-to-image diffusion model to manipulate a single keyframe in the video (Sec.~\ref{subsec:keyframe_editing}). 
Subsequently, we estimate the dense semantic correspondence between the input and edited keyframes for shape deformation. 
The shape deformation of the keyframe serves as the \emph{bridge} between input and output videos for per-frame deformation through the UV mapping and atlas. 
Our deformation module (Sec.~\ref{subsec:deformation}) transforms the UV map with the semantic correspondence to associate with the edits for each frame. 
To address the issues of unseen pixels from the single keyframe, we optimize the edited atlas and the deformation parameters guided by a pre-trained diffusion model with the input prompt (Sec.~\ref{subsec:optimization}).
% This process leads to a consistent video editings.
% Finally, a consistently edited video is achieved.
}
    \label{fig:overview}
\end{figure*}

Given an input video $\I^s_{1..N}$ and a text prompt, our proposed shape-aware video editing method produces a video $\I^t_{1..N}$ with appearance \emph{and} shape changes while preserving the motion in the input video. 
For maintaining temporal consistency, our method uses the pre-trained video decomposition method, NLA~\cite{kasten2021nla}, to acquire the canonical atlas layer $\I^s_A$ and the associated per-frame UV map $\uv^s_{A\to 1..N}$ per motion group. 
For simplicity, we assume a single moving object in an input video so that there are two atlases $\I^{s,FG}_{A}$ and $\I^{s,BG}_{A}$ for foreground and background contents, respectively. 
The edits in $\I^{s,FG}_{A}$ can be consistently transferred to each frame with UV mapping. 
To render the image $\I^s_j$ back, we use the $\uv^s_{A\to t}$ and an alpha map $\a^s_t$ to sample and blend:
\begin{equation}
\begin{split}
    \I^s_j &= \I_j^{s,FG} * \a^s_j + \I_j^{s,BG} * (1 - \a^s_j), \\
    \I_j^{s,g} &= \uv_{A\to j}^{s,g} \otimes \I_{A}^{s,g}, g \in \{FG, BG\},
\end{split}
\end{equation}
where $\otimes$ denotes the warping operation. 
Following our shape deformation introduction, we focus on the foreground atlas and will omit $FG$ from $\I^{s,FG}$ for simplicity.

We first select a single source keyframe $\I^{s}_k$ to pass into a text-driven image editing tool (\eg, Stable Diffusion~\cite{rombach2022sd}). 
The edits in target $\I^t_k$ will then be propagated to $\I^t_{1..N}$ through the atlas space with the mapping of $\uv^s_{A\to 1..N}$. 
Yet, the UV mapping cannot work when the edits involve \emph{shape changes} since $\uv^s_{A\to 1..N}$ are specifically for reconstructing the original shapes in the input video.
Hence, to associate the target shape correctly, we propose a UV deformation formulation (Sec.~\ref{subsec:deformation}) to transform each $\uv^s_{A\to j}$ into $\uv^t_{A\to j}$ according to the deformation between $(\I^s_k, \I^t_k)$. 
In other words, the keyframe deformation $\D^{s\to t}_k$ between $(\I^s_k, \I^t_k)$ serves as the \emph{bridge} between input and output videos for changing into the edited target shape while preserving the source motion in the input. 
Note that the edits and keyframe deformation $\D^{s\to t}_k$ alone are insufficient due to some unobserved areas from the viewpoint of image $\I^s_k$. 
Therefore, to acquire a complete and consistent editing result, we leverage a pre-trained diffusion model to optimize the editing appearance and deformation parameters in the atlas space in Sec.~\ref{subsec:optimization}. 
The process produces the final edited video $\I^t_{1..N}$ with desired object shape and appearance changes.

\subsection{Keyframe editing}
\label{subsec:keyframe_editing}
With the given text prompt, we edit a representative keyframe $\I^s_k$ (\eg, the middle frame of the video) by a pre-trained Stable Diffusion~\cite{rombach2022sd} to obtain target edited keyframe $\I^t_k$. 
Afterward, we leverage a pre-trained semantic correspondence model~\cite{truong2022probabilistic} to associate the correspondence between two different objects. 
The pixel-level semantic correspondence is the deformation that transforms the target shape in $\I^t_k$ to the source shape in $\I^s_k$. 
% We will elaborate on it in the next section.

\subsection{Deformation formulation}
\label{subsec:deformation}

With the estimated semantic correspondence, we can obtain the pixel-level \emph{shape deformation vectors}, $\D^{t\to s}_k \in \R^{H\times W\times 2}$. 
The target shape in $\I^t_k$ are then deformed into the source shapes in $\I^{s}_k$ via $\D^{t\to s}_k$:
\begin{equation}
    \I_{k}^{t\to s} = \D_k^{t\to s} \otimes \I_{k}^{t}.
\end{equation}

With the aid of $\D_k^{t\to s}$, the edited object can be back-projected to the atlas to form an edited atlas, $\I^{t\to s}_{A}$, by $\uv^s_{k\to A}$. 
Since it maintains the original shape, we cannot directly map the edited $\I^{t}_k$ to the atlas with $\uv^s_{k\to A}$.
\begin{figure*}
    \centering
    \includegraphics[width=\linewidth]{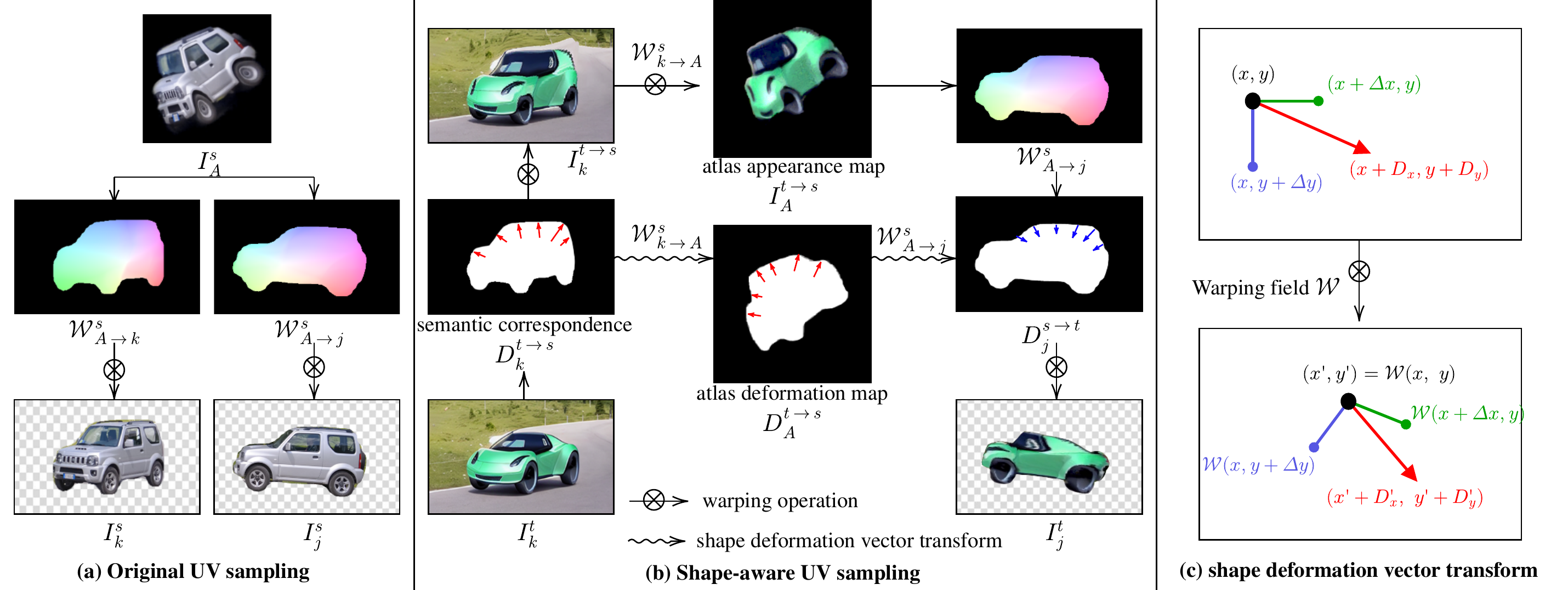}
\caption{\textbf{Deformation formulation.} 
Given the semantic correspondence between the input and edited keyframes, we map the edits back to the atlas via the original UV map (in the shape of the original atlas). 
Meanwhile, we transform the per-pixel deformation vectors into the atlas space with the same UV mapping field by (c). 
Consequently, the UV map samples the color and the deformation vectors onto each frame to deform the original UV map respecting the edited shape.
}
    \label{fig:deformation_initialization}
\end{figure*}

% Revised by Yi-Ting
Given the edited atlas $\I^{t\to s}_{A}$, the appearance edits can already be propagated to each frame with $\uv^s_{A\to 1..N}$ in source shapes. 
However, this needs improvement since our goal is to generate a new video with the target shape.
In addition to propagating the edited appearance via the atlas space, we spread the displacement vectors to each frame to obtain per-frame deformation by back projecting keyframe deformation $\D^{t\to s}_k$ into atlas space $A$ with $\uv^s_{k\to A}$ to get $\D^{t\to s}_A$. 
Yet, simply \emph{warping} into the new image space is insufficient as the coordinate system also got transformed by the warping operation. 
Therefore, we formulate a \emph{shape deformation vector transformation} matrix, $\M_{\uv}$, to handle the deformation vectors w.r.t. the original coordinate system by a warp field $\uv$:
\begin{equation}
    \D'(x', y')^T = \M_\uv D(x, y)^T,
\end{equation}
where $(x, y)$ and $(x', y')$ represent the corresponding pixels in the source and target images, respectively, by the warping field, $\uv$ (\ie, $(x', y') = \uv(x, y))$. 
For pixel-level deformation, we compute a per-pixel deformation vector $\M_{\uv}$ for each pixel $(x, y)$ by:
\begin{equation}
    \M_{\uv} = 
    \begin{bmatrix}
        \uv(x+\Delta x, y) - \uv(x, y) \\
        \uv(x, y+\Delta y) - \uv(x, y)
    \end{bmatrix}^T
    \begin{bmatrix}
        1 / \Delta x \\
        1 / \Delta y
    \end{bmatrix},
\end{equation}
where $\Delta x$ and $\Delta y$ denote small scalar shifts to form the local coordinate system in the source space. 
In practice, to avoid discrete sampling of warping, we use thin-plate spline~\cite{bookstein1989tps} to approximate the warping field smoothly. 
We illustrate the transformation of the shape deformation vector in Fig.~\ref{fig:deformation_initialization}c. 
With the transformation for the vector, we can obtain the corresponding deformation in the target warped space with the warp function $\uv$, which is the UV map in the atlas framework. 
Thus, the deformation map $\D^{t\to s}_k$ is propagated to each $I^t_j$ by:
\begin{equation}
\begin{split}
    \D^{t\to s}_{A} &= \M_{\uv^s_{k\to A}} \star (\uv^{s}_{k\to A} \otimes \D^{t\to s}_k) \\
    \D^{t\to s}_{j} &=  \M_{\uv^s_{A\to j}}\star (\uv^s_{A\to j} \otimes \D^{t\to s}_{A}),
\end{split}
\end{equation}
where $\star$ denotes the per-pixel matrix multiplication for the deformation map. 
Hence, we can deform the UV map $\uv^s_{A\to j}$ into $\uv^t_{A\to j}$ by $\uv^t_{A\to j} = \D^{s\to t}_{j} \otimes \uv^s_{A\to j}$.
Note that the alpha map for blending the target-shape object is also deformed in the same manner by $\a^t_j=\D^{s\to t}_{j}\otimes \a^s_j$.
Finally, the edited $\I^t_j$ with initial deformation on the foreground object can be obtained by:
\begin{equation}
\begin{split}
    \I^t_j = \uv^{t}_{A\to j} \otimes \I^{t\to s}_{A} * \a^t_j + \I^{BG}_A * (1 - \a^t_j).
\end{split}
\end{equation}

\subsection{Atlas optimization}
\label{subsec:optimization}
Through the deformation formulation in Sec.~\ref{subsec:deformation}, we can already obtain an edited video with the corresponding shape changes if the semantic correspondence, \ie, $D^{t\to s}_k$, is reliable. 
However, the estimated semantic correspondence is often inaccurate for shape deformation. 
As a result, it would yield distortions in some frames. 
Moreover, the edited atlas could be incomplete since it only acquires the editing pixels from the single edited keyframe so the unseen pixels from the keyframe are missing. 
Hence, these incomplete pixels produce visible artifacts in other frames. 

To address these issues, we utilize an additional atlas network $F_{\theta_A}$ and semantic correspondence network $F_{\theta_{SC}}$ to fill the unseen pixels and refine the noisy semantic correspondence via an optimization. 
Here, the atlas network $F_{\theta_A}$ takes the initial appearance and deformation of the foreground atlas $(I_A^{t\to s}, D_A^{t\to s})$ as input and outputs the \emph{refined} $(\tilde{\I}^{t\to s}_{A}, \tilde{\D}^{t\to s}_{A})$. 
Similarly, the semantic correspondence $\D^{\t\to s}_{k}$ is approximated by a thin-plate spline. 
We feed the control points into the semantic correspondence network $F_{\theta_{SC}}$ to obtain the refined $\tilde{\D}^{\t\to s}_{k}$.

We select several frames that capture different viewpoints for optimization. 
Our training of synthesizing the edited frames, $\I^t$, is guided by a pre-trained Vision-Language model with the target prompt. 
Inspired by DreamFusion~\cite{poole2022dreamfusion}, we leverage a pre-trained diffusion model~\cite{rombach2022sd} to provide pixel-level guidance by backpropagating the gradient of noise residual to the generated images (\emph{without} backpropagating through the U-Net model). 
Adding a noise $\e$ on $\I^t$ as the input, the pretrained diffusion UNet outputs a predicted noise $\hat{\e}$. 
The gradient of the noise residual $\hat{\e} - \e$ is backpropagated to update $\theta$:
\begin{equation}
    \nabla_{\theta}\L_{diff}(\I^t) \triangleq \mathbb{E}_{i,\e}[w(i)(\hat{\e}-\e)\frac{\partial\I^t}{\partial\theta}],
\end{equation}
where $i$ stands for the time step for the diffusion model and the parameter set $\theta = \{\theta_A, \theta_{SC}\}$. 
We update the unified information in the atlas space to maintain the temporal consistency of the editing appearance and deformation with only training on a few generated frames $\I^t$.

In addition to the guidance of the diffusion model on multiple frames, we also apply several constraints to the learning of the refinement networks, $F_{\theta_{A}}$ and $F_{\theta_{SC}}$,  to preserve the editing effects as in the target edited keyframe $I^t_k$. 
%For the updated atlas parameters $\tilde{\I}^{t\to s}_{A}$ and $\tilde{D}^{t\to s}_{A}$, we measure the error between the initial $\I^{t\to s}_{A}$ and $\D^{t\to s}_{A}$ by L1 loss to follow the target edits. 
To ensure that the deformation through the atlas can successfully reconstruct the original edited $\I^t_k$, the keyframe loss, $\L_{k}$, measures the error between the original $\I^t_k$ and the reconstructed $\tilde{\I}^t_k$ by L1 loss:
\begin{equation}
\L_{k} = |\tilde{\I}^t_k - \I^t_k|.
\end{equation}

Besides, we also apply a total variation loss to encourage the spatial smoothness of the refined appearance in the atlas. The atlas loss is as follows:
\begin{equation}
\begin{split}
    \L_{A} = \L_{tv}(\tilde{\I}^{t\to s}_{A}).
\end{split}
\end{equation}

During the optimization, we also refine the semantic correspondence $\tilde{\D}^{t\to s}_k$ of the keyframe pair. 
An ideal semantic correspondence matches semantically-similar pixels and perfectly transforms the target shape into the source shape. 
Therefore, we compute the errors of the deformed target and the source object masks, $\m^t_k$ and  $\m^s_k$:
\begin{equation}
    \L_{SC} = |(\tilde{\D}^{t\to s}_k \otimes \m^t_k) - \m^s_k|
\end{equation}

The total loss function $\L=\L_{diff}+\lambda_{k}\L_{k}+\lambda_A\L_{A}+\lambda_{SC}\L_{SC}$, $\lambda_k, \lambda_A, \lambda_{SC}=10^6, 10^3, 10^3$. 
The optimized parameters $\theta^*$ are then used to generate the final edited video $\I^{t*}_{1..N}$.

\topic{Implementation details.}

We implement our method in PyTorch. 
We follow the video configuration in NLA with the resolution of $768\times 432$. 
We use a thin-plate spline to inverse a warping field to prevent introducing holes by forward warping. 
The refinement networks, $F_{\theta_A}$ and $F_{\theta_{SC}}$ exploits the architecture of Text2LIVE~\cite{bar2022text2live} and TPS-STN~\cite{jaderberg2015stn}, respectively.
The optimization performs on 3 to 5 selected frames, including $I^t_1$, $I^t_k$, and $I^t_N$, for 600 to 1000 iterations. 
The optimization process takes 20 mins on a 24GB A5000 GPU. We further utilize an off-the-shelf super-resolution model~\cite{wang2021realesrgan} to obtain sharp details in the final edited atlases.

\begin{figure*}
    \centering
    
    \mpage{0.185}{\small Input}
    \mpage{0.185}{\small Ours}
    \mpage{0.185}{\small Multi-frame baseline}
    \mpage{0.185}{\small Single-frame baseline}
    \mpage{0.185}{\small Text2LIVE~\cite{bar2022text2live}} \\
    %{\includegraphics[width=0.195\linewidth]{figures/visual_comparison/blackswan/input/00006.png}} \hfill
    %{\includegraphics[width=0.195\linewidth]{figures/visual_comparison/blackswan/ours/00006.png}} \hfill
    %{\includegraphics[width=0.195\linewidth]{figures/visual_comparison/blackswan/multi/00006.png}} \hfill
    %{\includegraphics[width=0.195\linewidth]{figures/visual_comparison/blackswan/single/00006.png}} \hfill
    %{\includegraphics[width=0.195\linewidth]{figures/visual_comparison/blackswan/text2live/00006.png}} \\
    \frame{{\includegraphics[width=0.195\linewidth]{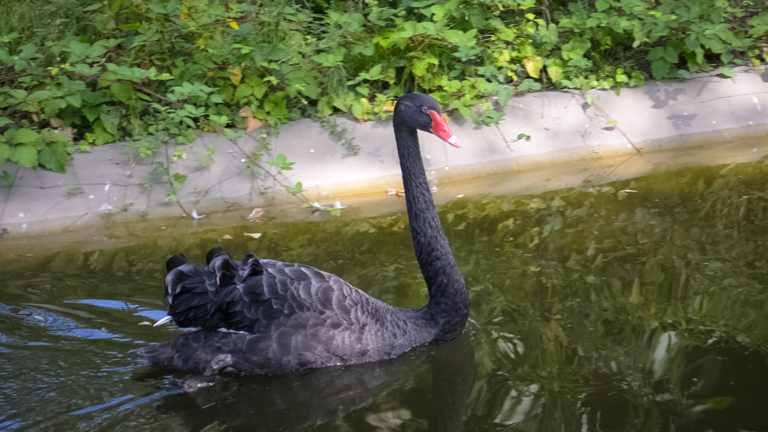}}} \hfill
    \frame{{\includegraphics[width=0.195\linewidth]{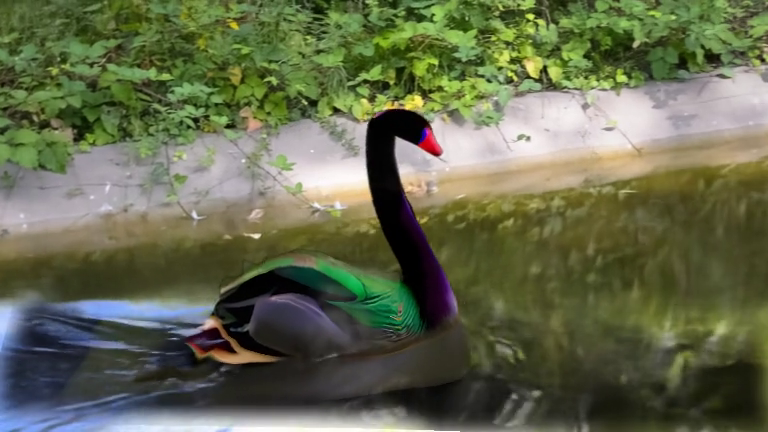}}} \hfill
    \frame{{\includegraphics[width=0.195\linewidth]{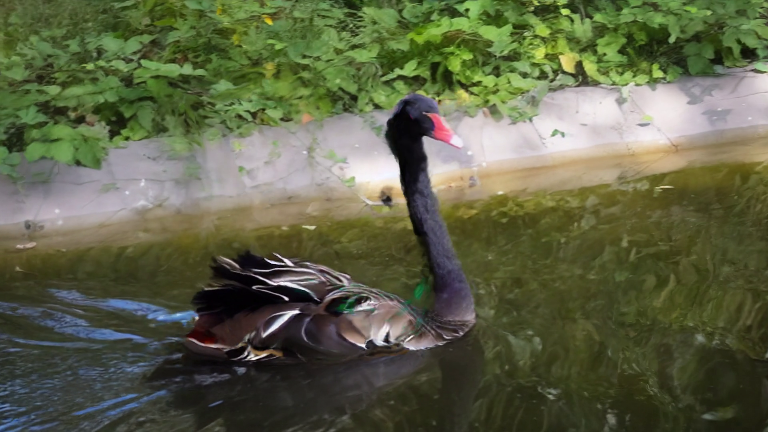}}} \hfill
    \frame{{\includegraphics[width=0.195\linewidth]{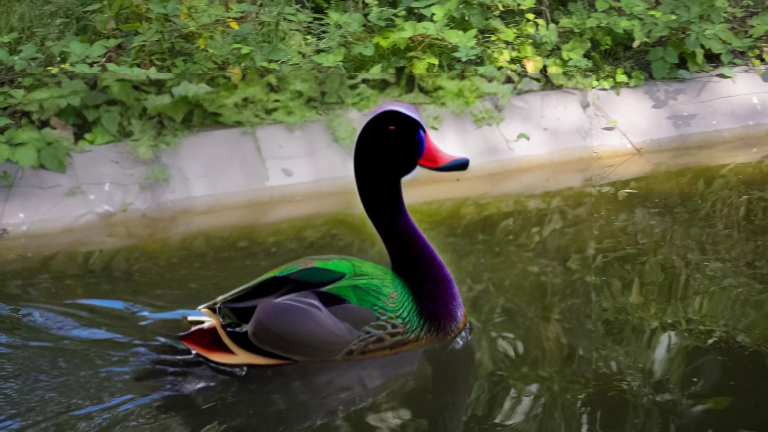}}} \hfill
    \frame{{\includegraphics[width=0.195\linewidth]{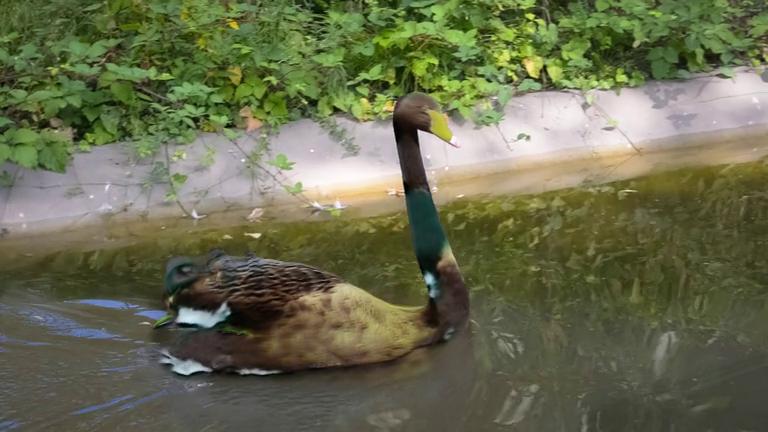}}} \\
    \vspace{0.05cm}
    \frame{{\includegraphics[width=0.195\linewidth]{figures/visual_comparison/blackswan/input/00043.png}}} \hfill
    \frame{{\includegraphics[width=0.195\linewidth]{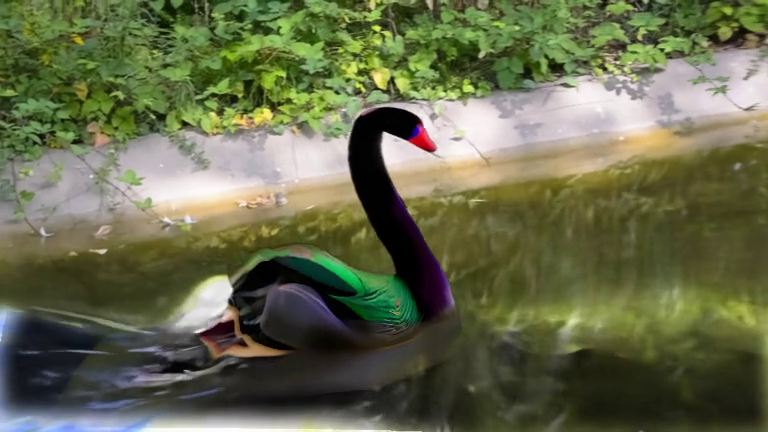}}} \hfill
    \frame{{\includegraphics[width=0.195\linewidth]{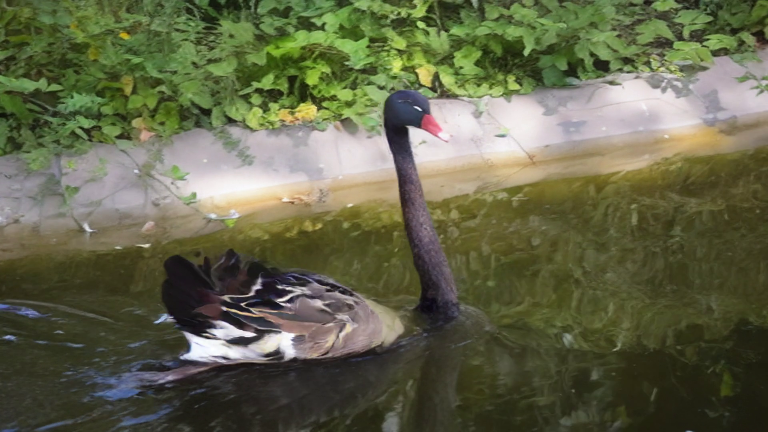}}} \hfill
    \frame{{\includegraphics[width=0.195\linewidth]{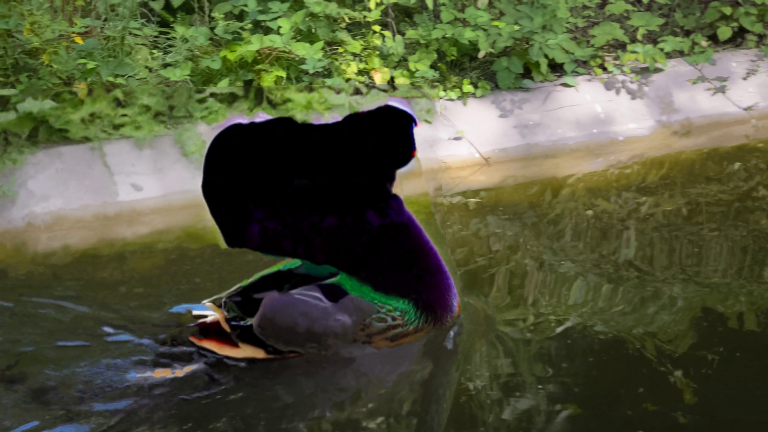}}} \hfill
    \frame{{\includegraphics[width=0.195\linewidth]{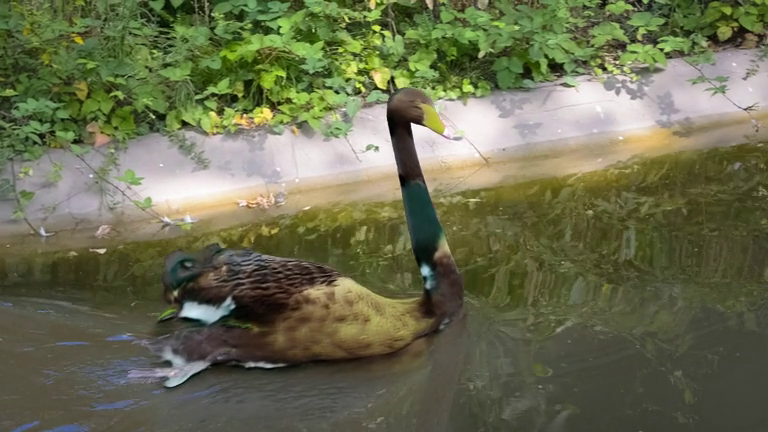}}} \\
    \vspace{0.05cm}
    \frame{{\includegraphics[width=0.195\linewidth]{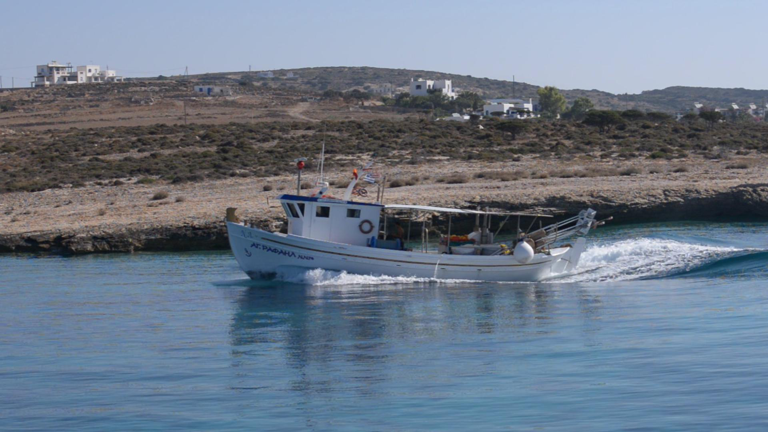}}} \hfill
    \frame{{\includegraphics[width=0.195\linewidth]{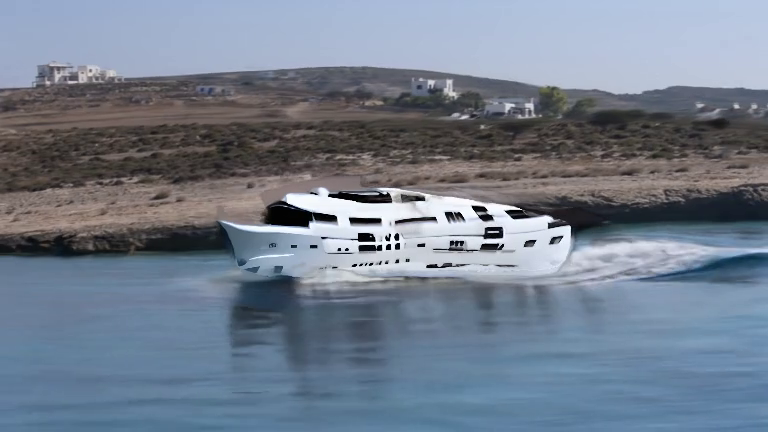}}} \hfill
    \frame{{\includegraphics[width=0.195\linewidth]{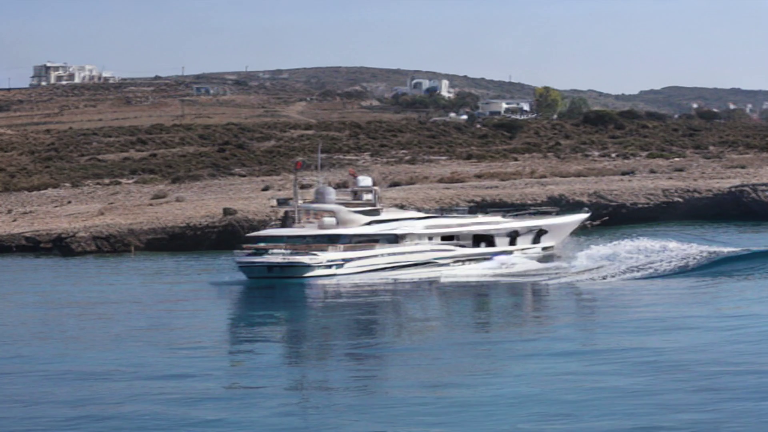}}} \hfill
    \frame{{\includegraphics[width=0.195\linewidth]{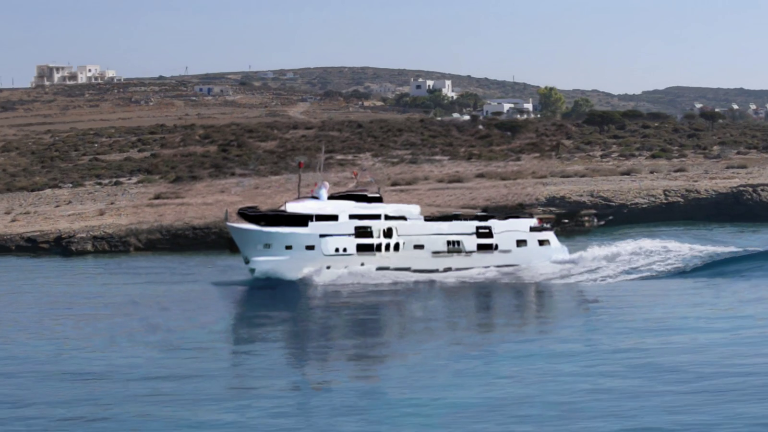}}} \hfill
    \frame{{\includegraphics[width=0.195\linewidth]{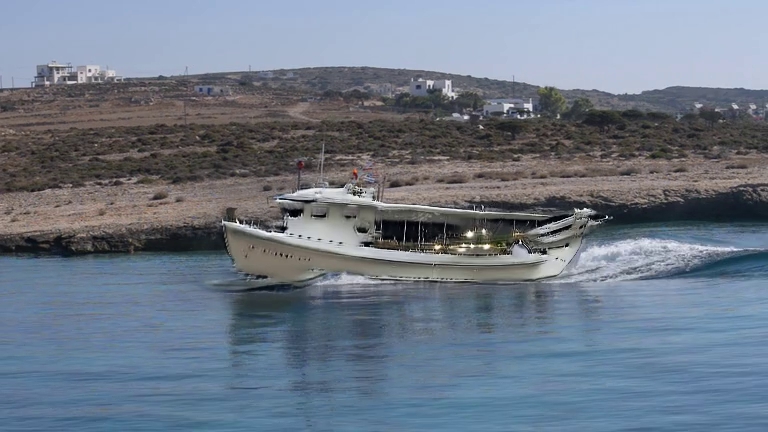}}} \\
    \vspace{0.05cm}
    %{\includegraphics[width=0.195\linewidth]{figures/visual_comparison/boat/input/00037.png}} \hfill
    %{\includegraphics[width=0.195\linewidth]{figures/visual_comparison/boat/ours/00037.png}} \hfill
    %{\includegraphics[width=0.195\linewidth]{figures/visual_comparison/boat/multi/00037.png}} \hfill
    %{\includegraphics[width=0.195\linewidth]{figures/visual_comparison/boat/single/00037.png}} \hfill
    %{\includegraphics[width=0.195\linewidth]{figures/visual_comparison/boat/text2live/00037.png}} \\
    \frame{{\includegraphics[width=0.195\linewidth]{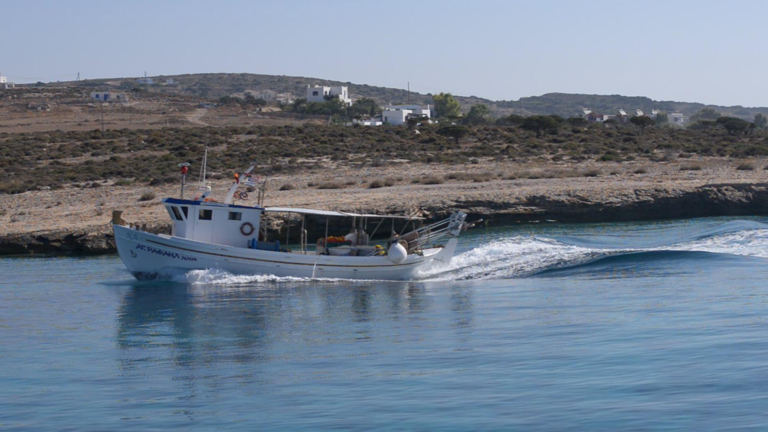}}} \hfill
    \frame{{\includegraphics[width=0.195\linewidth]{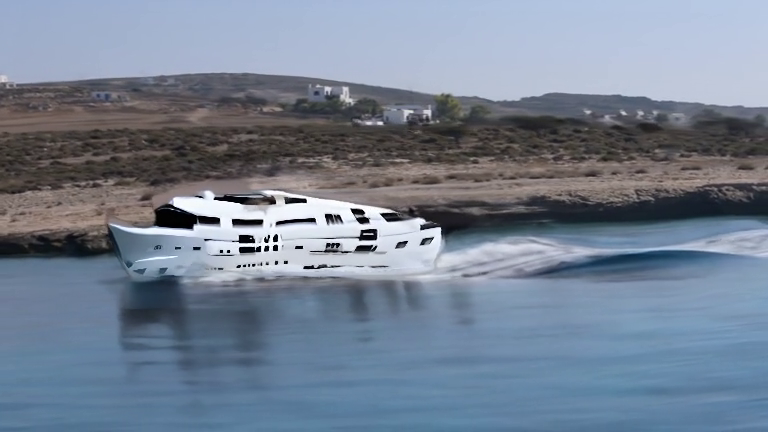}}} \hfill
    \frame{{\includegraphics[width=0.195\linewidth]{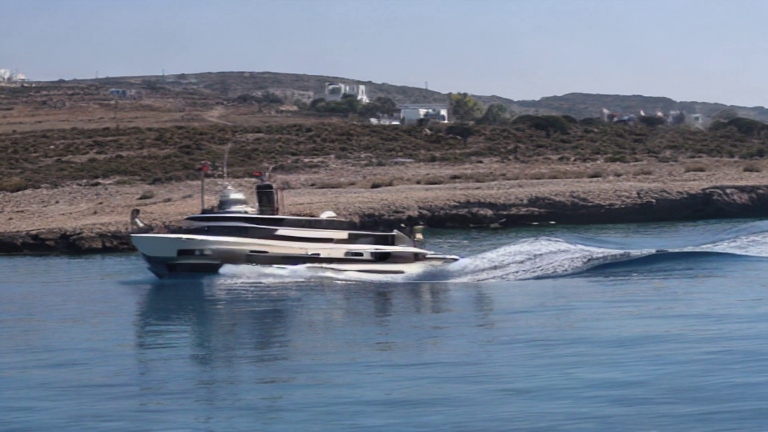}}} \hfill
    \frame{{\includegraphics[width=0.195\linewidth]{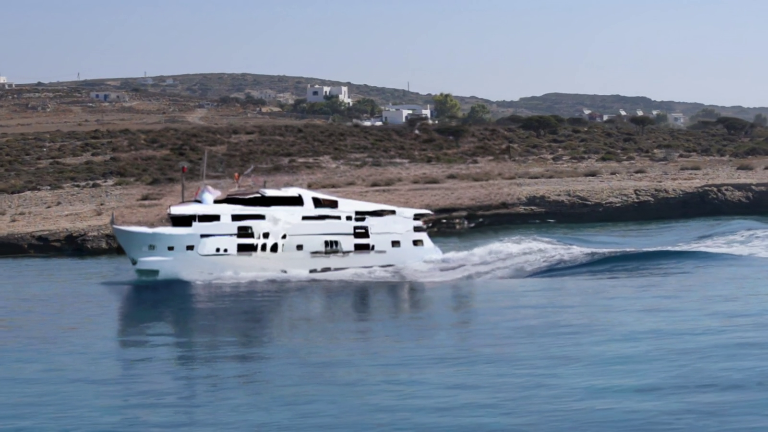}}} \hfill
    \frame{{\includegraphics[width=0.195\linewidth]{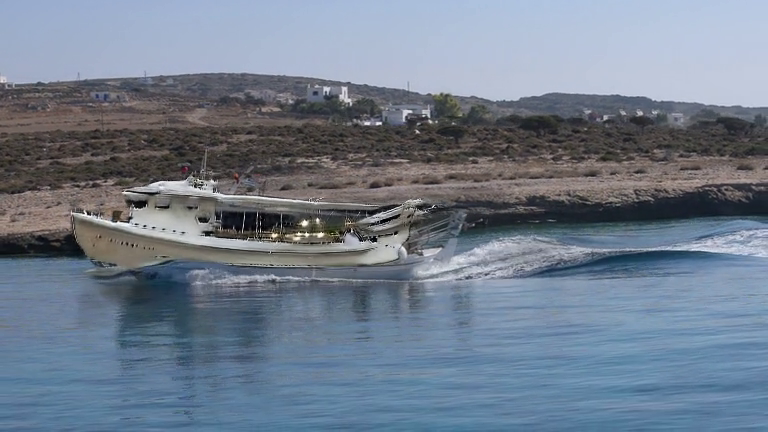}}} \\
    \vspace{0.05cm}
    \frame{{\includegraphics[width=0.195\linewidth]{figures/visual_comparison/libby/input/00003.png}}} \hfill
    \frame{{\includegraphics[width=0.195\linewidth]{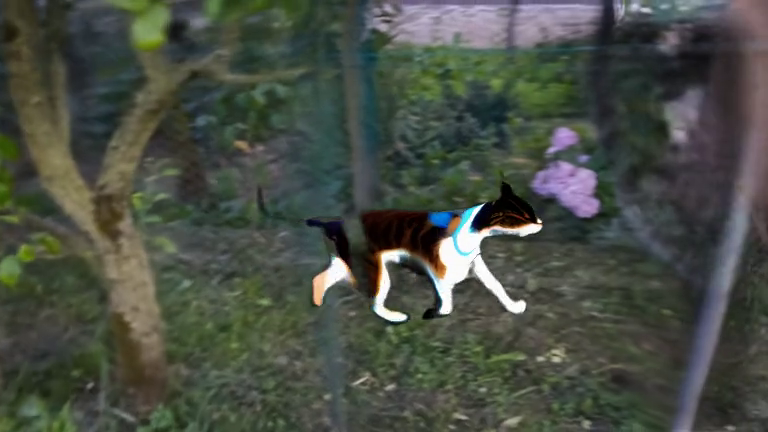}}} \hfill
    \frame{{\includegraphics[width=0.195\linewidth]{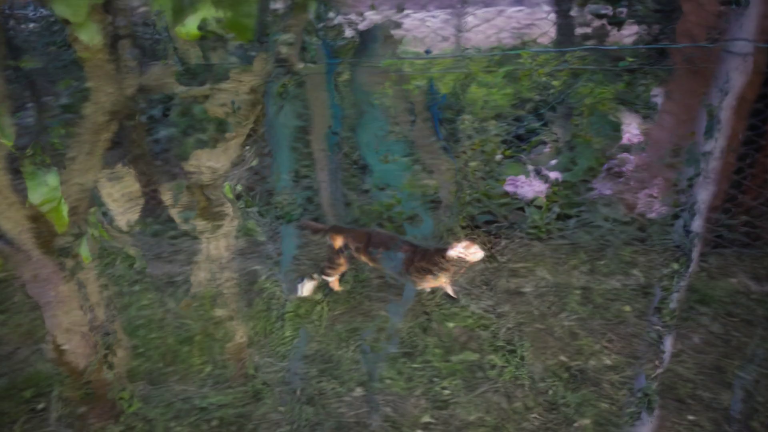}}} \hfill
    \frame{{\includegraphics[width=0.195\linewidth]{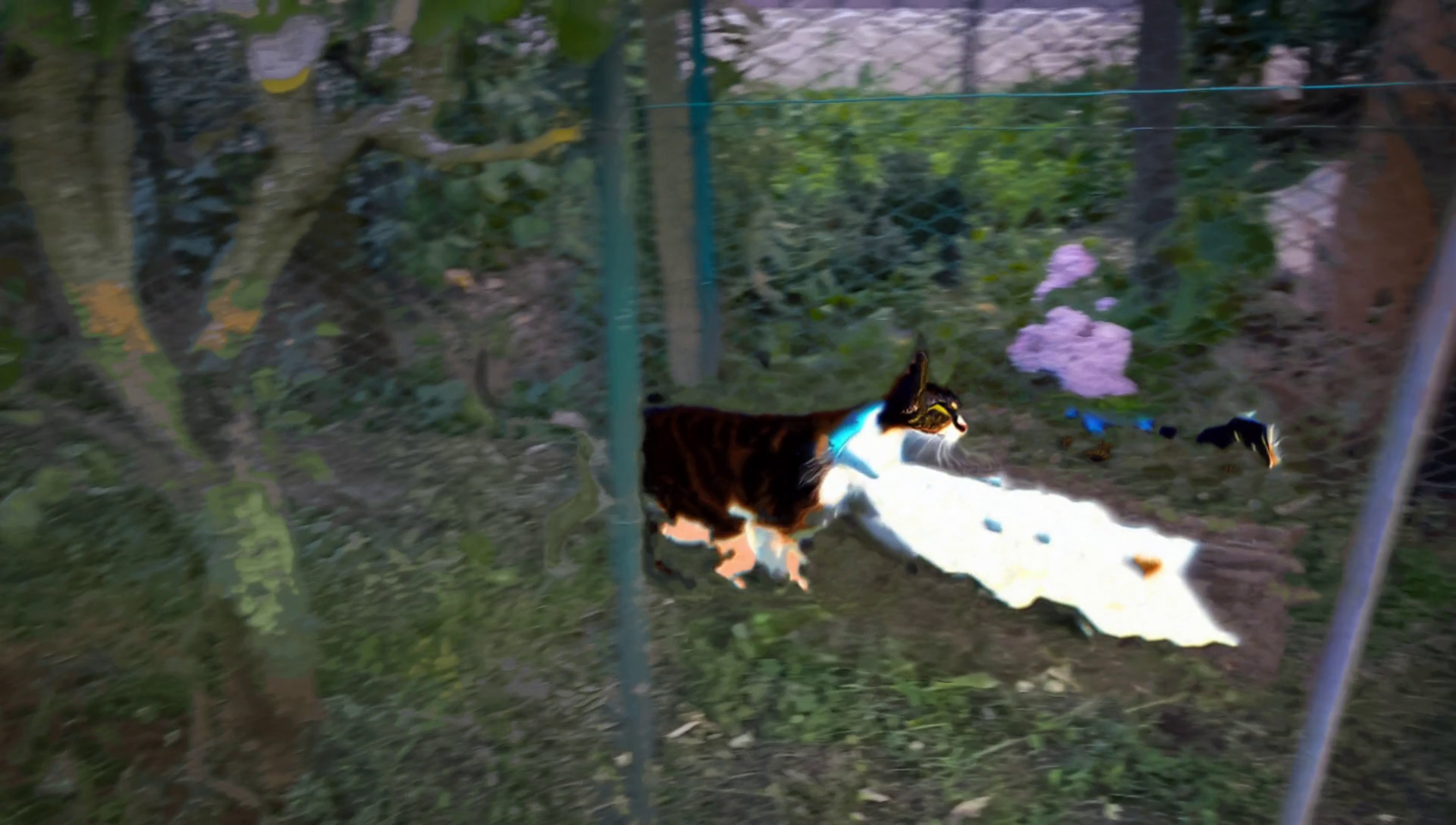}}} \hfill
    \frame{{\includegraphics[width=0.195\linewidth]{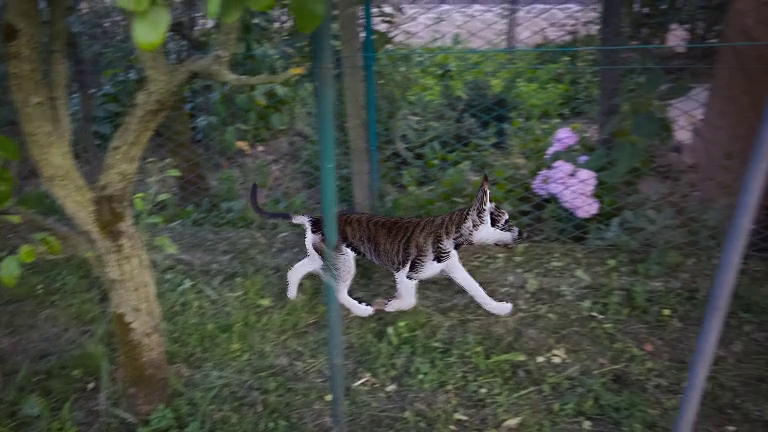}}} \\
    \vspace{0.05cm}
    \frame{{\includegraphics[width=0.195\linewidth]{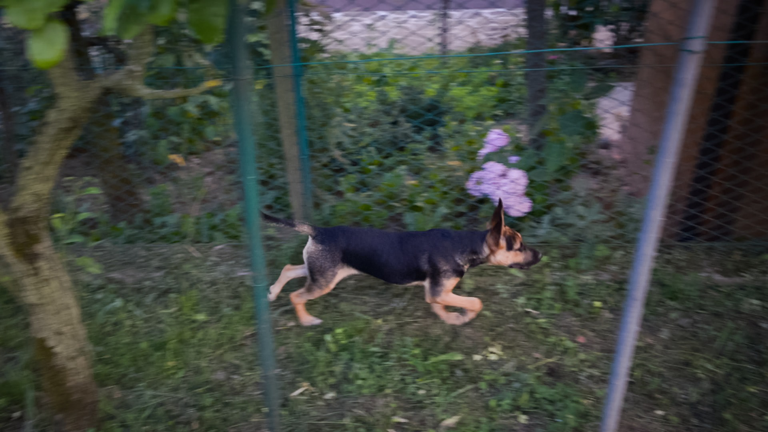}}} \hfill
    \frame{{\includegraphics[width=0.195\linewidth]{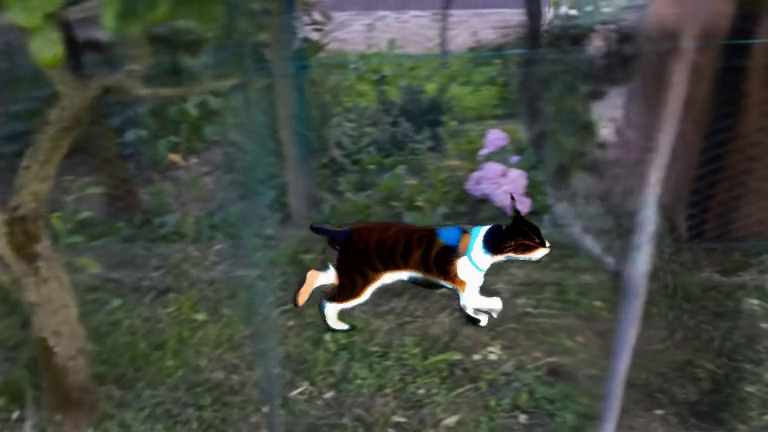}}} \hfill
    \frame{{\includegraphics[width=0.195\linewidth]{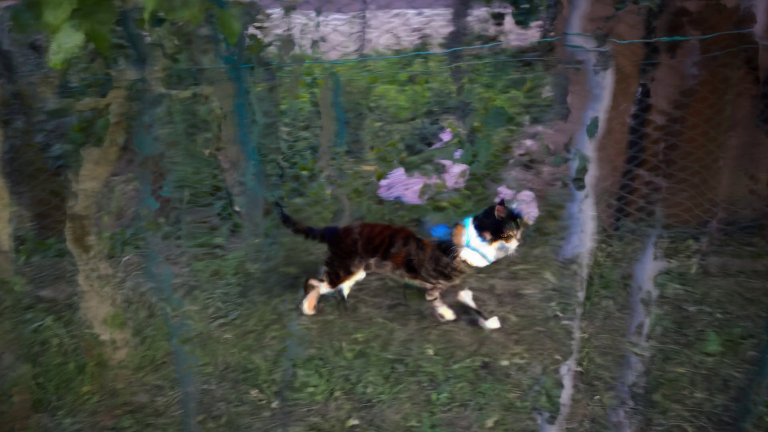}}} \hfill
    \frame{{\includegraphics[width=0.195\linewidth]{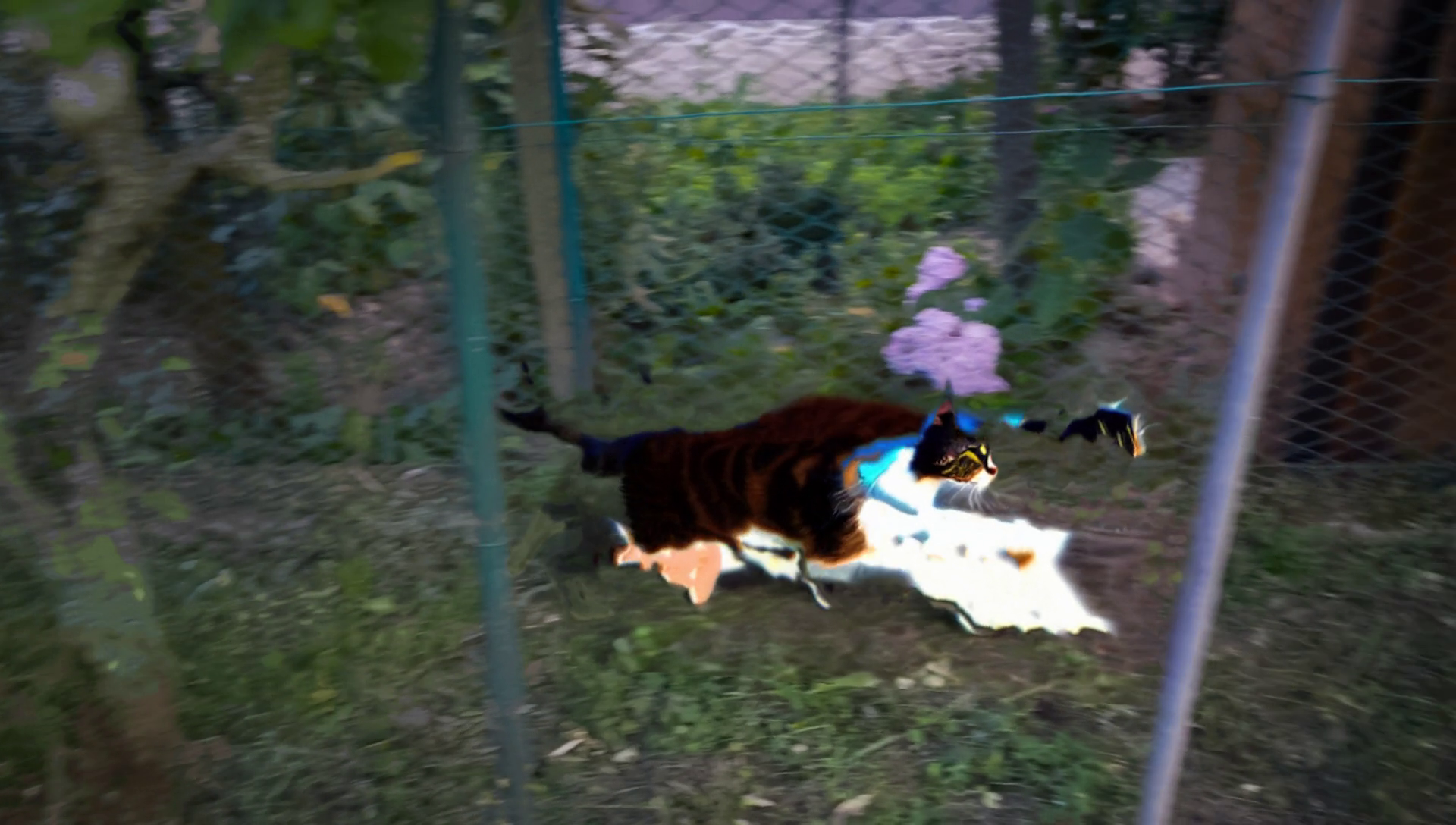}} }\hfill
    \frame{{\includegraphics[width=0.195\linewidth]{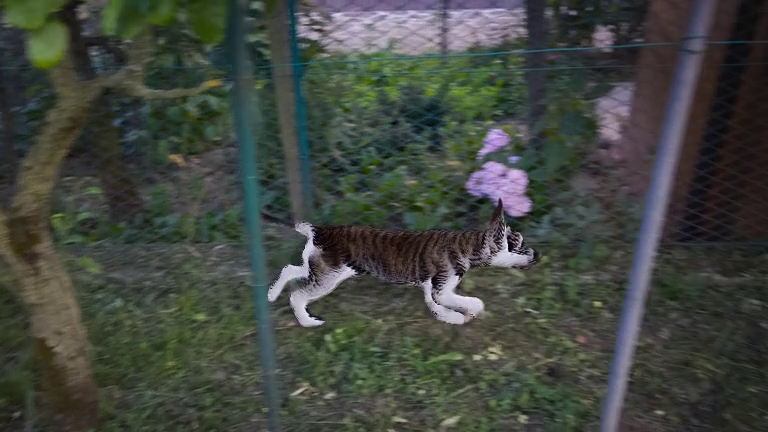}}} \\
    %{\includegraphics[width=0.195\linewidth]{figures/visual_comparison/libby/input/00009.png}} \hfill
    %{\includegraphics[width=0.195\linewidth]{figures/visual_comparison/libby/ours/00009.png}} \hfill
    %{\includegraphics[width=0.195\linewidth]{figures/visual_comparison/libby/multi/00009.png}} \hfill
    %{\includegraphics[width=0.195\linewidth]{figures/visual_comparison/libby/single/00009.png}} \hfill
    %{\includegraphics[width=0.195\linewidth]{figures/visual_comparison/libby/text2live/00009.png}} \\

\caption{\textbf{Visual comparison with baselines and SOTA.} 
We show three examples with edits of ``\texttt{blackswan} \textrightarrow \texttt{duck}'', ``\texttt{boat} \textrightarrow \texttt{yacht}'', and  ``\texttt{dog} \textrightarrow \texttt{cat}''. 
%We show two examples with edits of ``\texttt{blackswan} \textrightarrow \texttt{duck}'' and  ``\texttt{dog} \textrightarrow \texttt{cat}''. 
The multi-frame baseline shows inconsistency in the edited objects. 
The single-frame method suffers from the incomplete flow motion of the source object shape and thus could not propagate the edits properly. 
Text2LIVE demonstrates consistent appearance editing corresponding to the target edits. Nevertheless, the shape remains the same as the original object.
% it limits by the fixed UV mapping so that the resulting shapes remain the same as the original object. 
In contrast, our proposed method outperforms the compared methods with consistent and plausible appearance and shape editing. 
% Zoom in for the best view.
}
    \label{fig:visual_comparison}
\end{figure*}

\begin{figure*}
    \centering    
    %\frame{{\includegraphics[width=0.195\linewidth]{figures/ablation/00003_crop/input_crop.png}}} \hfill
    %\frame{{\includegraphics[width=0.195\linewidth]{figures/ablation/00003_crop/wo_sc_crop.png}}} \hfill
    %\frame{{\includegraphics[width=0.195\linewidth]{figures/ablation/00003_crop/wo_deform_crop.png}}} \hfill
    %\frame{{\includegraphics[width=0.195\linewidth]{figures/ablation/00003_crop/wo_optim_crop.png}}} \hfill
    %\frame{{\includegraphics[width=0.195\linewidth]{figures/ablation/00003_crop/full_crop.png}}} \\
    \frame{{\includegraphics[width=0.195\linewidth]{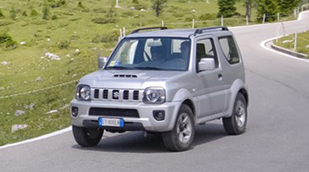}}} \hfill
    \frame{{\includegraphics[width=0.195\linewidth]{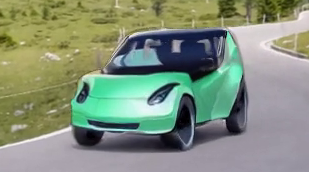}}} \hfill
    \frame{{\includegraphics[width=0.195\linewidth]{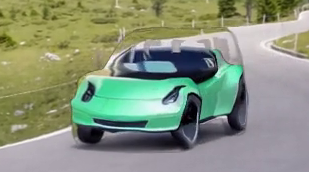}}} \hfill
    \frame{{\includegraphics[width=0.195\linewidth]{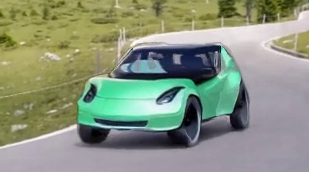}}} \hfill
    \frame{{\includegraphics[width=0.195\linewidth]{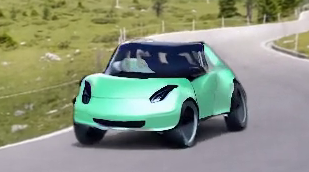}}} \\
    \vspace{0.05cm}
    \frame{{\includegraphics[width=0.195\linewidth]{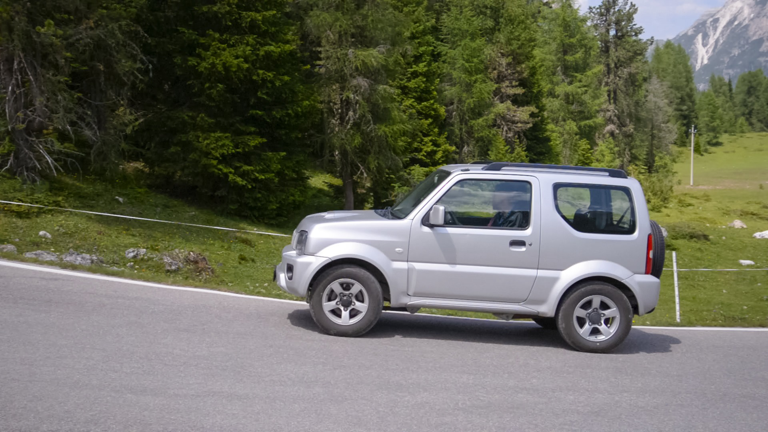}}} \hfill
    \frame{{\includegraphics[width=0.195\linewidth]{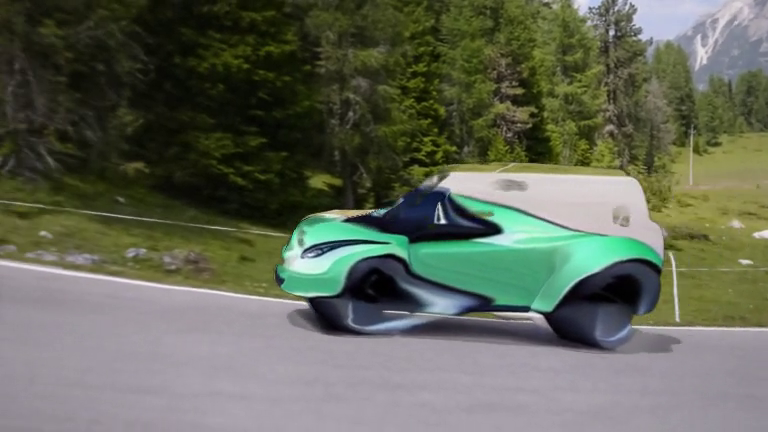}}} \hfill
    \frame{{\includegraphics[width=0.195\linewidth]{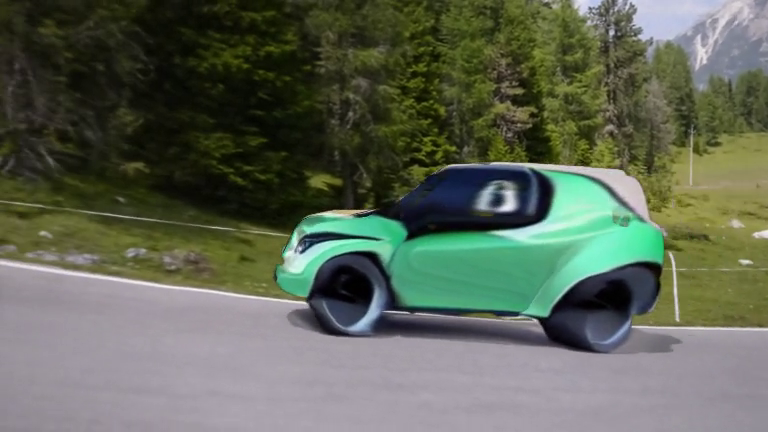}}} \hfill
    \frame{{\includegraphics[width=0.195\linewidth]{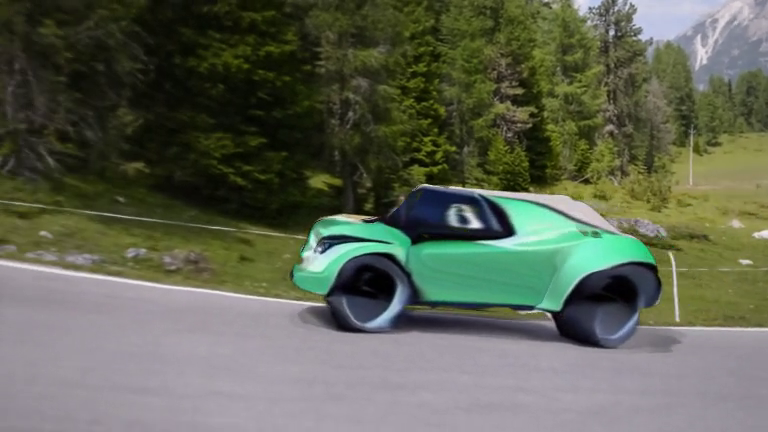}}} \hfill
    \frame{{\includegraphics[width=0.195\linewidth]{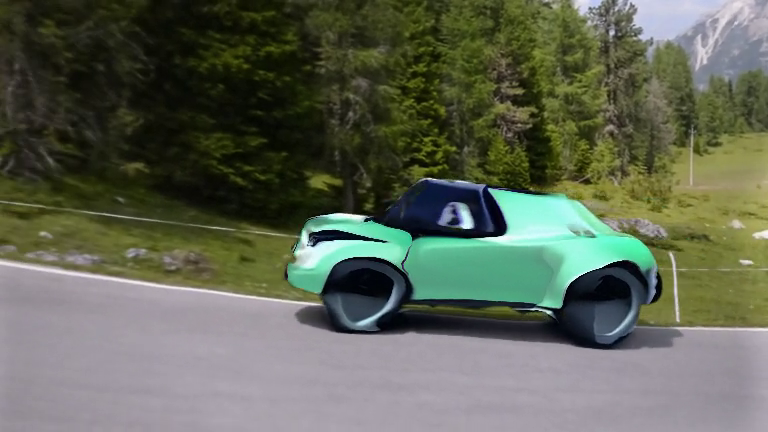}}} \\
    \mpage{0.185}{\small Input}
    \mpage{0.185}{\small (a) fixed NLA}
    \mpage{0.185}{\small (b) w/ semantic corres.}
    \mpage{0.185}{\small (c) w/ UV deformation}
    \mpage{0.185}{\small (d) w/ optimization (full)} \\
\caption{\textbf{Ablation study.} 
We study the effects of removing the deformation and optimization components. 
(a) Editing with fixed NLA UV mapping. (b) Using a semantic correspondence with fixed UV, the edits are mapped to the atlas properly but still remains the original shapes in results. 
(c) With deformation initialization (Sec.~\ref{subsec:deformation}), the NLA UV maps are deformed to restore the target shape. 
(d) With further atlas optimization (Sec.~\ref{subsec:optimization}), the incomplete pixels in edited atlas and distortion (in car's roof and back wheel) are refined.}
\label{fig:ablation}
\end{figure*}

\begin{figure*}
    \centering
    \frame{{\includegraphics[width=0.245\linewidth]{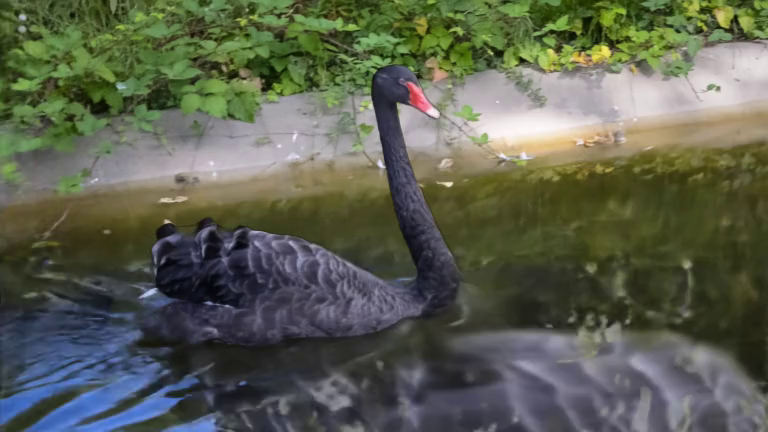}}} \hfill
    \frame{{\includegraphics[width=0.245\linewidth]{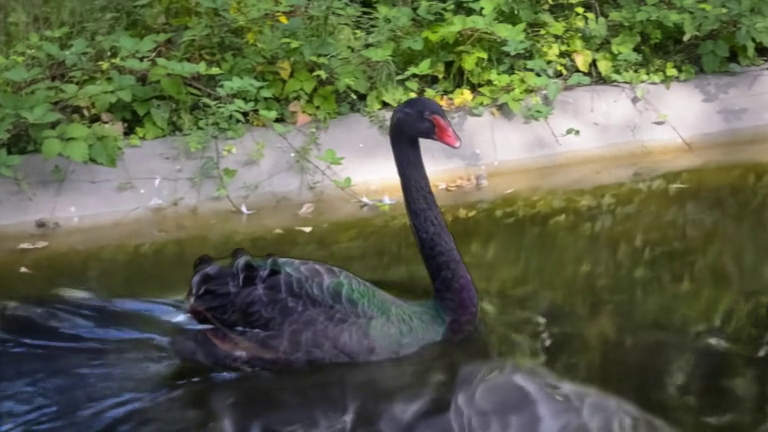}}} \hfill
    \frame{{\includegraphics[width=0.245\linewidth]{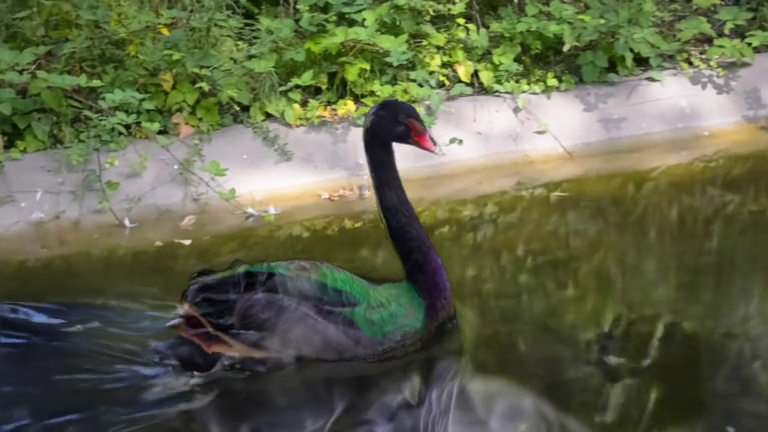}}} \hfill
    \frame{{\includegraphics[width=0.245\linewidth]{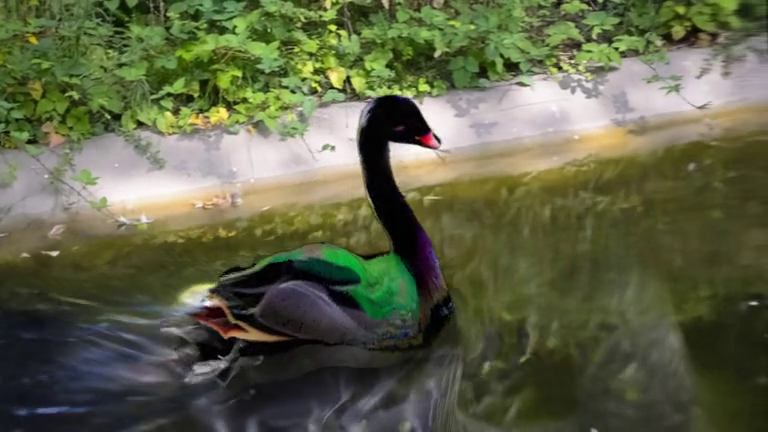}}} \\
    \vspace{0.05cm}
    %\frame{{\includegraphics[width=0.245\linewidth]{figures/application/motorbike2/00007.png}}} \hfill
    %\frame{{\includegraphics[width=0.245\linewidth]{figures/application/motorbike2/00019.png}}} \hfill
    %\frame{{\includegraphics[width=0.245\linewidth]{figures/application/motorbike2/00028.png}}} \hfill
    %\frame{{\includegraphics[width=0.245\linewidth]{figures/application/motorbike2/00036.png}}} \\
    \frame{{\includegraphics[width=0.245\linewidth]{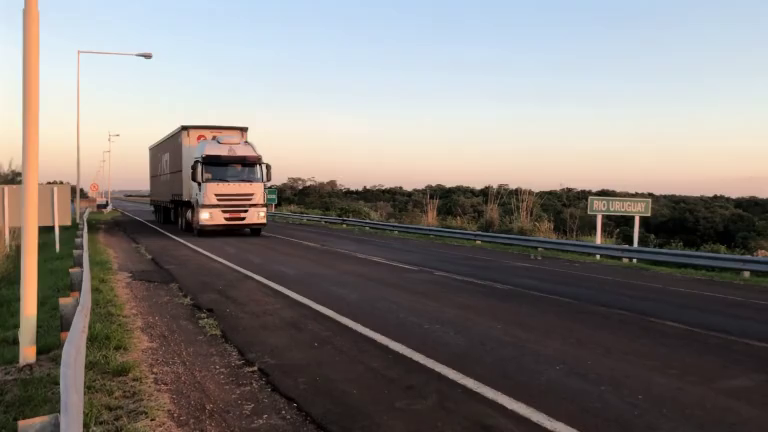}}} \hfill
    \frame{{\includegraphics[width=0.245\linewidth]{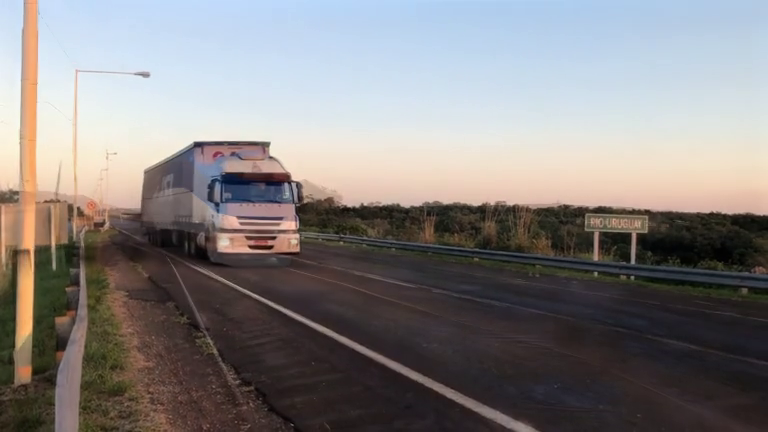}}} \hfill
    \frame{{\includegraphics[width=0.245\linewidth]{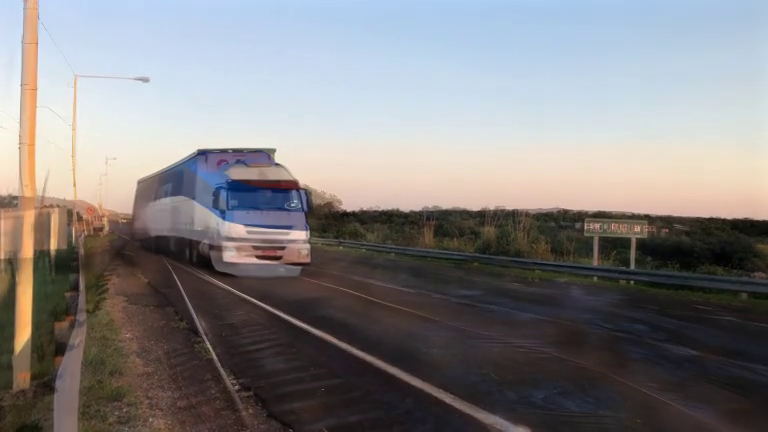}}} \hfill
    \frame{{\includegraphics[width=0.245\linewidth]{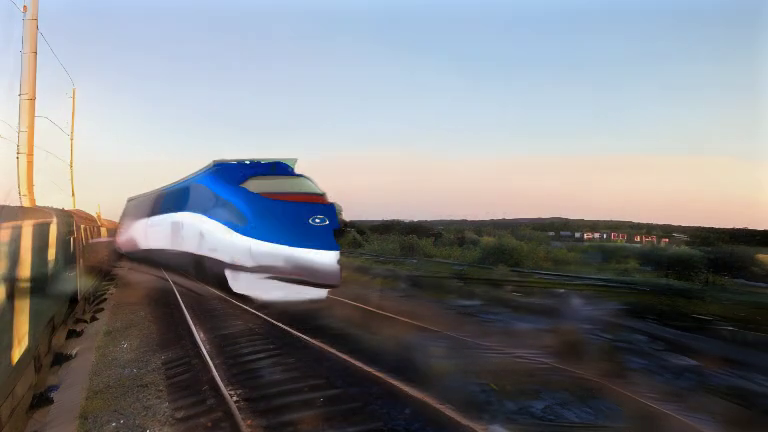}}} \\
    %\vspace{0.1cm}
    %{\includegraphics[width=0.245\linewidth]{figures/application/SUV/00010.png}} \hfill
    %{\includegraphics[width=0.245\linewidth]{figures/application/SUV/00025.png}} \hfill
    %{\includegraphics[width=0.245\linewidth]{figures/application/SUV/00032.png}} \hfill
    %{\includegraphics[width=0.245\linewidth]{figures/application/SUV/00044.png}} \\
    \vspace{-0.2cm}\caption{
    \textbf{Shape-aware interpolation.} 
    Our methods allow interpolation between two shapes by simply interpolating the atlas deformation maps. 
    The examples demonstrate the gradual changes from source objects to edited objects over the time. 
    % Best view in our demo video.
    }\vspace{-0.3cm}
    \label{fig:application}
\end{figure*}

\section{Experimental Results}
\label{sec:result}

Here we show sample editing results in the paper. 
We include additional video results in the supplementary material. 
We will make our source code and editing results publicly available to foster reproducibility.

\subsection{Experimental Setup}
\label{sec:setup}

\topic{Dataset.} 
We select several videos from DAVIS~\cite{pont2017davis}. 
Each video contains a moving object in 50 to 70 frames. 
We edit each video with a prompt that describes a target object with a different shape from the original one.

\topic{Compared methods.}
We compare our results with SOTA and several baseline methods. 
For fair comparisons, all the baseline methods use the same image editing method, Stable Diffusion~\cite{rombach2022sd}.

\noindent\textbullet~\textbf{Multi-frame baseline}: Multiple keyframes in a video are edited individually. 
The nearby edited keyframes temporally interpolate the remaining frames with FILM~\cite{reda2022film}.

\noindent\textbullet~\textbf{Single-frame baseline}: 
We extract a single keyframe from a video to be edited. 
The edited information is then propagated to each frame with EbSynth~\cite{jamrivska2019ebsynth}.
    
\noindent\textbullet~\textbf{Text2LIVE}~\cite{bar2022text2live}: 
The SOTA text-driven editing method with NLA. 
Note that it utilizes a structure loss to preserve the original shape. 
We compare the official Text2LIVE in this section and show the comparison of removing structure loss in our supplementary material.

\subsection{Visual Comparison}
\label{sec:visual_comparison}
We show a visual comparison with the baseline methods and Text2LIVE in Fig.~\ref{fig:visual_comparison}. 
% We encourage readers to see the visual comparison in our supplementary video.
In the first example with ``\texttt{blackswan}\textrightarrow\texttt{duck}'', the multi-frame baseline shows inconsistent editing in different frames. 
The single-frame baseline suffers from inaccurate frame motion and thus yields distortion during propagation. 
Text2LIVE shows a promising target appearance with temporal consistency but cannot change the shape that matches the target object. 
In contrast, our method provides the desired appearance \emph{and} consistent shape editing. 
In the second example with ``\texttt{boat}\textrightarrow\texttt{yacht}'', the single-frame baseline shows an inconsistent shape since the frame propagation relies on the frame motion of the source shape. 
Consequently, it cannot propagate the edited pixels correctly in a different shape. 
In the third example with ``\texttt{dog}\textrightarrow\texttt{cat}'', the input video contains a non-rigid motion moving object. 
It poses further challenges for multi- and single-frame baselines. 
Again, Text2LIVE demonstrates plausible \texttt{cat} appearance while remaining in the source \texttt{dog} shape. Our shape-aware method maintains the object motion and manipulates the texture and shape corresponding to the desired editing.

\subsection{Ablation Study}
\label{sec:ablation}
We conduct an ablation study in Fig.~\ref{fig:ablation} to validate the effectiveness of the UV deformation and atlas optimization. 
With fixed NLA UV mapping, the shape edits in the keyframe cannot be adequately transformed through the atlas to each frame (Fig.~\ref{fig:ablation}a). 
Therefore, by adding a keyframe semantic correspondence to deform the target into the source shape, the fixed UV maps the edits correctly into the atlas but remains source shapes in the edited frames (Fig.~\ref{fig:ablation}b). 
To restore the target shape, our deformation module deforms the UV maps by the semantic correspondence (Fig.~\ref{fig:ablation}c). 
However, the unseen pixels and inaccurate correspondence yield artifacts in different views (\eg, in the car's roof and back wheel). We refine the edited atlas and deformation with the atlas optimization (Fig.~\ref{fig:ablation}d).

\subsection{Application}
\label{sec:application}
We present an application of shape-aware interpolation in Fig.~\ref{fig:application}. 
Through interpolating the deformation maps, the object shape can be easily interpolated \emph{without} additional frame interpolation methods. 
Similarly, we can interpolate atlas textures. 
Note that we directly apply image editing on the background atlas since it can be treated as a natural panorama image (shown in Fig.~\ref{fig:overview}). 
However, the foreground atlas is an unwrapped object texture, which is unnatural for general pre-trained editing models. 
Therefore, we edit the video frame and map it back to the atlas. 
This approach is more general and allows users to use their chosen images for video editing. 
% We will release our codes. already say it up-front

\subsection{Limitations}
\label{sec:limitations}

\begin{figure}
    \centering
    %{\mfigure{0.235}{figures/failure/input_00000_crop.png}} \hfill 
    \mpage{0.02}{\raisebox{50pt}{\rotatebox{90}{Input}}}
    \frame{{\includegraphics[width=0.475\linewidth]{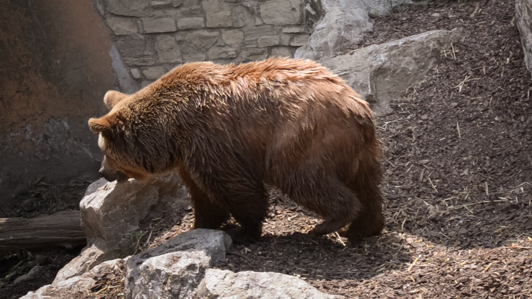}}}
    %{\mfigure{0.235}{figures/failure/input_00015_crop.png}} \hfill 
    \frame{{\includegraphics[width=0.475\linewidth]{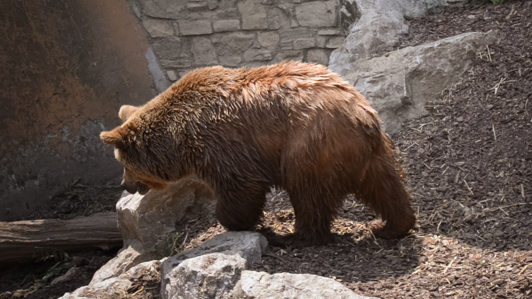}}} \\
    \vspace{-1.1cm}
    %{\mfigure{0.235}{figures/failure/output_00000_crop.png}} \hfill 
    \mpage{0.02}{\raisebox{50pt}{\rotatebox{90}{Edit}}}
    \frame{{\includegraphics[width=0.475\linewidth]{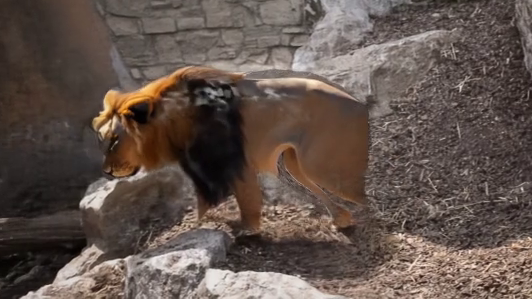}}}
    %{\mfigure{0.235}{figures/failure/output_00015_crop.png}} \hfill 
    \frame{{\includegraphics[width=0.475\linewidth]{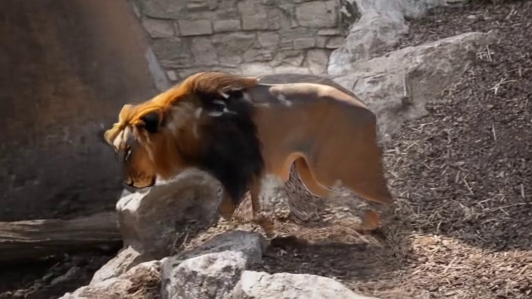}}} \\
    \vspace{-1.1cm}
    \caption{\textbf{Limitations.} 
    We visualize a failure example (\texttt{bear} \textrightarrow \texttt{lion}). 
    The inaccurate NLA mapping in the motion of crossing hind legs yields distortion in the edited result.
    }\vspace{-0.3cm}
    \label{fig:limitations}
\end{figure}

\begin{figure}
    \centering
    {\includegraphics[width=0.49\linewidth]{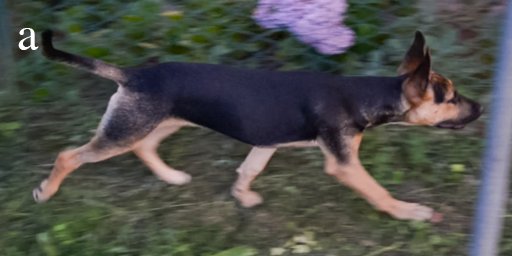}} 
    {\includegraphics[width=0.49\linewidth]{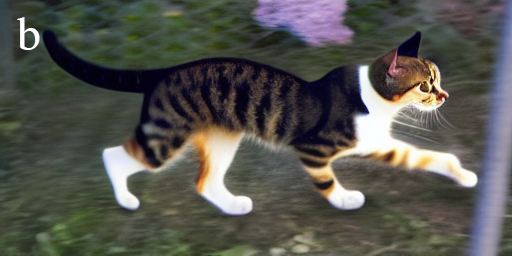}} \\
    {\includegraphics[width=0.49\linewidth]{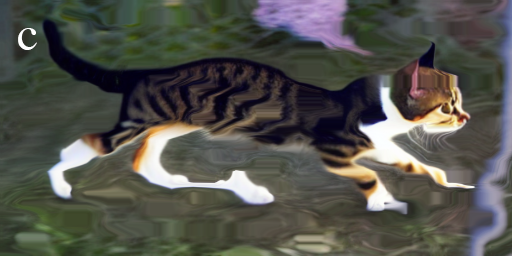}} 
    {\includegraphics[width=0.49\linewidth]{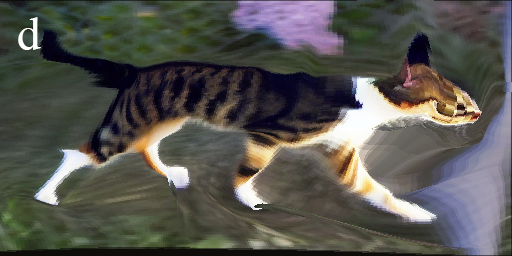}} \\
    \vspace{-0.2cm}\caption{\textbf{User-guided correspondence.} 
Associating two different objects remains challenging even for the SOTA semantic correspondence methods. 
For a pair of source (a) and target (b), the severe false matching can be corrected by users' manual warping for better results.
}\vspace{-0.3cm}
    \label{fig:manual}
\end{figure}

% Our approach consists of the following limitations. 
Our method strictly relies on the \emph{many-to-one} mapping from individual frames to a unified atlas. 
However, NLA may fail to get the ideal mapping in challenging scenarios with complex motions.
Therefore, we observe artifacts in the erroneous mapping regions (\eg, the motion of hind legs shown in Fig.~\ref{fig:limitations}). 
In addition, it remains difficult to build semantic correspondence between two different objects.
While the atlas optimization can improve noisy correspondences, poor semantic correspondence initialization would hinder the optimization. 
We show that user manual correction (in Fig.~\ref{fig:manual}) can lead to better video editing results.
% Thus, we can get the correspondence from the manual warping by users (in Fig.~\ref{fig:manual}) to encourage better video editing results.

\section{Conclusions}
\label{sec:conclusions}

We have presented a shape-aware text-driven video editing method. 
We tackle the limitation of appearance-only manipulation in existing methods. 
We propose a deformation formulation using layered video representation to transform the mapping field corresponding to the target shape edits. 
% By obtaining the deformation between a pair of source keyframes and the edited keyframe, the deformation information could be propagated to each frame through the atlases.
We further refine the unseen regions by utilizing the guidance from a pre-trained text-to-image diffusion model. 
Our method facilitates a variety of shape and texture editing applications.

\topic{Societal impacts.} 
Our work proposes a tool for enabling creative video editing applications.
Nevertheless, similar to many image/video synthesis applications, care should be taken to prevent misuse or malicious use of such techniques. 
We will release our code under a similar license as Stable Diffusion that focuses on ethical and legal use.\footnote{\url{https://github.com/CompVis/stable-diffusion/blob/main/Stable_Diffusion_v1_Model_Card.md}}
% as many image/video synthesis applications, 
% for saving users' manual efforts in complicated video editing in many contexts. 
% Nevertheless, editing real data into fake one is a common double-edged sword issue. 
% Malicious parties may misuse these powerful techniques, such as the well-known DeepFake. 
% Therefore, these issues should be kept concerned in the related research.

%[Figure 1] Teaser
%- input frames (4 frames)
%- polar bear 
%- lion 
%-- 3 x 4 column

%[Figure 2] Limitations of existing work
%- Text2Live (4 frames) -> cannot change shape
%- per-frame -> flicker
%- per-frame + EbSyn -> artifacts

% https://www.microsoft.com/en-us/research/wp-content/uploads/2017/01/structure_completion_small.pdf

%[Figure 3] Overview

%[Figure 4] Initialize 

%[Figure 5] Visual comparison

%[Figure 6] Application
%- Same video, but three different edits

%[Figure 7] Ablation
%- w and w/o initialization 
%- w and w/o optimization 
%- w and w/o atlas inpainting

%[Figure 8] User-guided semantic correspondence

%[Figure 9] User study

%[Figure 10] Failure cases

%%%%%%%%% REFERENCES
{\small
%\bibliographystyle{ieee_fullname}
%\bibliography{egbib}

}

\end{document}